\documentclass[letterpaper, 10 pt, conference]{ieeeconf}  

\IEEEoverridecommandlockouts                              

\usepackage{amsmath} 
\usepackage{xcolor}
\usepackage{amssymb}  
\usepackage{soul}
\usepackage{graphicx}
\usepackage{subcaption} 
\usepackage{caption}
\usepackage[font=small,labelfont=bf]{caption}
\usepackage{hyperref}
\usepackage{multirow}
\usepackage{booktabs}

\soulregister{\ref}{1}

\title{\LARGE \bf
DuLoc: Life-Long Dual-Layer Localization in Changing\\ and Dynamic Expansive Scenarios
}

\author{Haoxuan Jiang, Peicong Qian, Yusen Xie, Xiaocong Li, Ming Liu, and Jun Ma, \textit{Senior Member, IEEE}
  \thanks{This work was supported by the Guangdong provincial project under Grant 2023QN10Z006. \textit{(Corresponding author: Jun Ma.)}}
\thanks{Haoxuan Jiang and Yusen Xie are with Robotics and Autonomous Systems Thrust, The Hong Kong University of Science and Technology (Guangzhou), 
        Guangzhou 511453, China (e-mail: hjiangax@connect.hkust-gz.edu.cn; yxie827@connect.hk-gz.edu.cn).}%
\thanks{Peicong Qian and Ming Liu are with Shenzhen Unity Drive Innovation Technology Co., Ltd.,
        Shenzhen 518063, China (e-mail: epsilonjohn9527@gmail.com; liu.ming.prc@gmail.com).}%
\thanks{Xiaocong Li is with the College of Information Science and Technology, Eastern
Institute of Technology, Ningbo, Ningbo 315200, China (e-mail: xiaocongli@eitech.edu.cn). }
\thanks{Jun Ma is with the Robotics and Autonomous Systems Thrust, The Hong Kong University of Science and Technology (Guangzhou), Guangzhou 511453, China, and also with the Division of Emerging Interdisciplinary Areas, The Hong Kong University of Science and Technology, Hong Kong SAR, China (e-mail: jun.ma@ust.hk).} 
}

\begin{document}

\maketitle

\thispagestyle{empty}
\pagestyle{empty}

\begin{abstract}

LiDAR-based localization serves as a critical component in autonomous systems, yet existing approaches face persistent challenges in balancing repeatability, accuracy, and environmental adaptability. Traditional point cloud registration methods relying solely on offline maps often exhibit limited robustness against long-term environmental changes, leading to localization drift and reliability degradation in dynamic real-world scenarios. 
To address these challenges, this paper proposes DuLoc, a robust and accurate localization method that tightly couples LiDAR-inertial odometry with offline map-based localization, incorporating a constant-velocity motion model to mitigate outlier noise in real-world scenarios. Specifically, we develop a LiDAR-based localization framework that seamlessly integrates a prior global map with dynamic real-time local maps, enabling robust localization in unbounded and changing environments. Extensive real-world experiments in ultra unbounded port that 
involve 2,856 hours of operational data across 32 Intelligent Guided Vehicles (IGVs) are conducted and reported in this study. The results attained demonstrate that our system outperforms other state-of-the-art LiDAR localization systems in large-scale changing outdoor environments.

\end{abstract}

\section{INTRODUCTION}
High-precision life-long localization in large-scale environments faces fundamental challenges across various autonomous systems \cite{5940562,10428583,10287946,10535355}. While Real-Time Kinematic (RTK) localization has been widely adopted, its reliability is inherently limited by satellite signal availability and near metallic structures, with typical accuracy degradation from centimeters to meters in such scenarios. Map-based localization \cite{9636501, 9982153} often encounters challenges due to environmental changes, leading to unstable and incorrect performance. These limitations drive the development of sensor-based localization approaches. Initial efforts focused on single-sensor solutions (e.g., LiDAR~\cite{koide2024glim} or cameras~\cite{ORBSLAM3_TRO}), but their susceptibility to environmental changes and perceptual degradation in feature-deprived areas proved inadequate for large-scale deployments~\cite{9788021,yin2024survey,10456930}.
The evolution to multi-sensor fusion systems \cite{9372856, 9304577, 9636655} address some shortcomings by combining complementary sensing modalities. However, these approaches still struggle with long-term consistency due to cumulative errors in odometry estimation and the absence of global constraints. Even state-of-the-art fusion methods exhibit gradual drift in kilometer-scale operations~\cite{9788021,yin2024survey,10456930}, particularly in dynamic environments where moving objects distort perception observations. This fundamental limitation underscores the need for persistent global references in large-scale localization.

To address these issues, in this paper, we propose a novel tightly-coupled framework that incorporates LiDAR-IMU odometry, integrates global offline map constraints, and leverages the constant velocity (CV) model, specifically designed to achieve accurate and robust localization in large-scale industrial environments.
Firstly, we propose a tightly-coupled LiDAR-Inertial-CV odometry backbone based on an iterated Error-State Kalman Filter (iESKF) that ensures robust short-term tracking. Then, we design a dual-map architecture combining a static prior map with a dynamic local map, enabling simultaneous global consistency and local adaptability. 
Finally, we validate our approach in one of the most demanding real-world environments: automated port operations. 
Essentially, modern container ports present extreme localization challenges~\cite{:/content/books/9789213584569}, including ultra large-scale unbounded scenes with persistent GPS-denial areas and severe multipath effects caused by dense metallic obstructions. The frequent cargo handling, vehicle movement, and machinery operations further intensify the highly dynamic nature of these settings, rendering precise localization exceptionally difficult.
In addition, the challenges of mapping in such large areas are compounded by high mapping costs, difficulty in maintenance, high environmental repetitiveness, and the stringent accuracy requirements for localization. In our experiments, 2,856 hours of operational data across 32 Intelligent Guided Vehicles (IGVs) are involved. Our system demonstrates centimeter-level accuracy (mean: 8.3\,cm, $\sigma$: 4.1\,cm) while maintaining 99.98$\%$ availability, outperforming existing LiDAR-SLAM baselines \cite{9304577, 9636655} that failed catastrophically within 30 minutes of operation due to misalignment of the map or odometry drift.

The main contributions of our work are summarized as follows:
\begin{itemize}
\item We present a highly efficient dual-map LiDAR localization framework integrated into a multi-sensor tightly-coupled localization system, specifically designed for real-world large-scale, dynamic, and rapidly changing environments. 

\item We import an optimized Kalman gain computation method to improve computational efficiency and enhance the accuracy of point cloud feature alignment during the filter measurement update process.

\item We propose a constant velocity (CV) model operating at up to 100 Hz, ensuring stable and continuous pose prediction even in case of sensor failures.

\item Our algorithm is deployed and validated in a port covering an area of approximately one million square meters, providing accurate and stable localization support for 32 IGVs; and this underscores the effectiveness of the proposed method in real-world settings.
\end{itemize}

\section{RELATED WORK}

\subsection{LiDAR-Based SLAM}

LiDAR-based SLAM methods are widely used for map construction and self-localization. LeGO-LOAM \cite{8594299} and F-LOAM \cite{9636655} both utilize ground planes for feature extraction and matching, with F-LOAM introducing a two-stage distortion compensation method. However, these methods struggle with handling dynamic environments and feature redundancy. SROM \cite{9304577} uses a two-layer approach to estimate rotation and translation, refining the results with point-to-plane ICP \cite{121791, 924423}, though it can be computationally expensive. PFilter \cite{9981566}, an extension of F-LOAM, filters invalid features to improve accuracy but may still suffer from high computational overhead in large-scale scenarios. FEVO-LOAM \cite{9866843} optimizes feature extraction with enhanced ground segmentation and curvature definitions, yet it may face limitations in complex urban environments with varied terrain. RF-LOAM \cite{10288340} uses FA-RANSAC to remove dynamic objects, but its effectiveness diminishes with extreme environmental changes. WiCRF2 \cite{10197305} enhances motion observability and minimizes redundancy, but it may struggle in highly dynamic or unstructured environments. The method in \cite{10632209} employs angle-based feature extraction and voxel-based feature matching but may be sensitive to LiDAR viewpoint variations. CDP-LOAM \cite{10637771} introduces clustering-directed points for attitude estimation, yet it is computationally heavy. Light-LOAM \cite{10439642} offers a two-stage correspondence strategy for reliable registration but may not scale well in large-scale dynamic environments.

\subsection{Map-Based LiDAR Localization}
Recent map-based LiDAR localization methods focus on balancing accuracy, efficiency, and adaptability. LOL \cite{9197450} reduces cumulative drift via geometric place recognition but struggles with partially similar scenes, suggesting dynamic map updates. DLL \cite{9636501} optimizes point-to-map distances without feature extraction, yet its computational intensity demands hierarchical optimization for scalability. For dynamic environments, ROLL \cite{9982153} activates temporary mapping during global matching failures, though latency issues necessitate adaptive triggering, while DMLL \cite{10285531} mitigates alignment errors through differential constraints but requires probabilistic error modeling to address complex uncertainties. Efficiency-driven approaches (Thakur \cite{10552773} and Block Map \cite{feng2024blockmapbasedlocalizationlargescaleenvironment}) leverage lightweight maps and block switching for large-scale applications, yet risk accuracy loss or transition misalignments, calling for multi-resolution maps and boundary refinement.

\subsection{Multi-Sensor Fused Localization}

Multi-sensor fusion methods combine LiDAR with inertial measurements to address motion distortion and environmental dynamics. Tightly-coupled frameworks dominate recent advancements, where LINS \cite{9197567} and LIO-EKF \cite{10610667} utilize iterated/extended Kalman filters with adaptive error compensation, though their computational load escalates in large-scale environments. Meanwhile, LIO-SAM \cite{9341176} enhances real-time performance through factor graphs and IMU pre-integration, while the FAST-LIO series \cite{9372856,9697912} innovates with state-dependent Kalman gain and raw point-to-map registration (FAST-LIO2 \cite{9697912}), significantly reducing feature extraction overhead at the cost of degraded accuracy during aggressive maneuvers. For high-dynamic scenarios, Point-LIO \cite{Point-LIO} introduces stochastic kinematic modeling to handle IMU saturation, yet its efficacy remains constrained by inertial sensor quality.


\section{METHODOLOGY}

\begin{figure}[t]
\centering
\includegraphics[width=\linewidth]{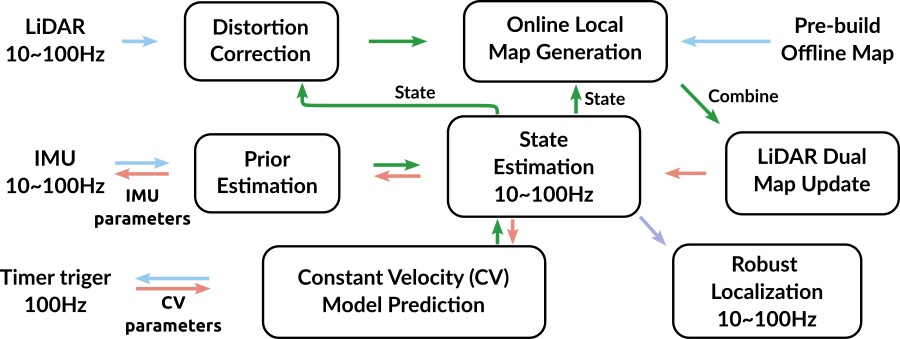}
\caption{The overall architecture of our proposed system, based on the iESKF filter, tightly integrates LiDAR, IMU data and constant velocity model for high-precision localization. The \textcolor[RGB]{0,0,255}{\textit{blue}} arrows indicate data input, the \textcolor[RGB]{41,154,63}{\textit{green}} arrows represent the forward process, the \textcolor[RGB]{232,138,122}{\textit{orange}} arrows depict the constraint and covariance propagation, and the \textcolor[RGB]{168,170,225}{\textit{purple}} arrows signify the output of the localization results.}
\label{Fig.1}
\vspace{-1 em}
\end{figure}

As illustrated in Fig.\ref{Fig.1}, our framework comprises several key modules: the system model in Sec.~\ref{sec:system_decri}, constant velocity prediction model in Sec.~\ref{sec:cvmodel}, dual-map update strategy in Sec.~\ref{sec:lidar_dual_update}, IMU correction model in Sec.~\ref{sec:imu_correction}, and additional implementation details including post-processing and LiDAR motion compensation in Sec.~\ref{sec:details_imple}.

\subsection{System Description}
\label{sec:system_decri}
Firstly, we define the state vector \textbf{x} of our system as
\begin{equation}
\label{eq:attn_G}
\resizebox{\linewidth}{!}{
\(
\mathbf{x} \triangleq \begin{bmatrix}
    ^{W}\mathbf{R}_{I} &
    ^{W}\mathbf{p}_{I} &
    ^{W}\mathbf{v}_{I} &
    \boldsymbol{\omega} &
    \mathbf{a} &
    \mathbf{b}_{\boldsymbol{\omega}} &
    \mathbf{b}_{\mathbf{a}} &
    ^{W}\!\mathbf{g} &
    ^{I}\mathbf{R}_{L} &
    ^{I}\mathbf{p}_{L}
\end{bmatrix}
\)
}
\end{equation}
where \( ^{W}\mathbf{R}_I \), \( ^{W}\mathbf{p}_I \), and \( ^{W}\mathbf{v}_I \) denote the IMU rotation, position, and velocity in the world frame $W$. \( \mathbf{b}_{\boldsymbol{\omega}} \) and \( \mathbf{b}_{\mathbf{a}} \) are the IMU biases of angular velocity $\omega$ and linear acceleration $\mathbf{a}$. \( ^{W}\!\mathbf{g} \) is the gravity vector in the world frame $W$. \( ^{I}\mathbf{R}_{L} \) and \( ^{I}\mathbf{p}_{L} \) are the extrinsic parameters between IMU frame and LiDAR frame.

Inertial Measurement Unit (IMU) provides motion information over short periods. When combined with inputs from other sensors and advanced algorithms, it achieves impressive accuracy within a tightly-coupled multi-sensor fusion framework.
The IMU noise \textbf{w} is defined as:
\begin{equation}
\mathbf{w} \triangleq \begin{bmatrix}
    \mathbf{n}_{\boldsymbol{\omega}} &
    \mathbf{n}_{\mathbf{a}} &
    \mathbf{n}_{\mathbf{b} \boldsymbol{\omega}} &
    \mathbf{n}_{\mathbf{b a}}
\end{bmatrix}
\end{equation}
where \( \boldsymbol{\omega} \) and \( \mathbf{a} \) are the IMU angular velocity and linear acceleration. \( \mathbf{n}_{\boldsymbol{\omega}} \) and \( \mathbf{n}_{\mathbf{a}} \) denote the measurement noise of  \( \boldsymbol{\omega} \) and \( \mathbf{a} \). \( \mathbf{n}_{\mathbf{b} \boldsymbol{\omega}} \) and \( \mathbf{n}_{\mathbf{b a}} \) are random walk process noises.

In our system, the state transition model at the sampling period \(\Delta t\) is defined as
\begin{equation}
\mathbf{x}_{i+1}=\mathbf{x}_{i} \boxplus \left(\Delta t \mathbf{f}\left(\mathbf{x}_{i}, \mathbf{w}_{i}\right)\right) \tag{3}
\end{equation}
where the symbol \( \boxplus \), as introduced in \cite{9372856}, represents the state transition of the system in the Lie algebra space. The function \textbf{f} in forward process can be derived as
\begin{equation}
\resizebox{0.8\linewidth}{!}{
\(
\mathbf{f}(\mathbf{x}, \mathbf{w}) \triangleq \begin{bmatrix}
    \boldsymbol{\omega}-\mathbf{b}_{\boldsymbol{\omega}} \\
    ^{W}\mathbf{v}_{I}+\frac{1}{2}\left(^{W}\mathbf{R}_{I}\left(\mathbf{a}-\mathbf{b}_{\mathbf{a}}\right)+^{W}\!\!\mathbf{g}\right) \Delta t \\
    ^{W}\mathbf{R}_{I}\left(\mathbf{a}-\mathbf{b}_{\mathbf{a}}\right)+^{W}\!\!\mathbf{g} \\
    \mathbf{n}_{\boldsymbol{\omega}} \\
    \mathbf{n}_{\mathbf{a}} \\
    \mathbf{n}_{\mathbf{b} \boldsymbol{\omega}} \\
    \mathbf{n}_{\mathbf{b a}} \\
    \mathbf{0}_{3 \times 1} \\
    \mathbf{0}_{3 \times 1} \\
    \mathbf{0}_{3 \times 1}
\end{bmatrix}
\)
}
\end{equation}

\subsection{Constant Velocity Model Prediction}
\label{sec:cvmodel}

The CV model is simple, robust, and reliable in environments with IMU data interruptions or sensor failures. Unlike the constant acceleration model (which risks instability), or machine-learning models (which depend heavily on data quality and training), the CV model avoids complexity and ensures stability. 
In the case of sensor failures, such as data interruptions from the IMU, we introduce a 100 Hz timer and switch to a CV model for prediction. This method ensures that even in the absence of IMU data, the localization system does not get stuck waiting for data synchronization or immediately diverge. The system assumes that the linear and angular velocity states remain constant between two consecutive LiDAR frames, allowing the entire positioning pipeline to degrade into a pure LiDAR-based localization method using the CV model. The prediction model, as applied during the filter prediction phase, can be expressed as:
\begin{equation}
\mathbf{x}_{i+1} = \mathbf{x}_{i} \boxplus \left(\Delta t \mathbf{f}\left(\mathbf{x}_{i}, \mathbf{0}\right)\right)
\end{equation}
where the covariance propagation follows a process similar to that of the iESKF algorithm.




\subsection{LiDAR Dual-Map Updates}
\label{sec:lidar_dual_update}

Compared to directly preforming point cloud registration with a prior map, we transform the feature alignment process into the filter measurement update process within the iESKF framework based on point-to-plane constraints in the map \( \mathcal{M} \):
$$
^\mathcal{M}\mathbf{R}_{j} \!\! \left(\!\mathbf{x}_{i}, \!^{L_{i}}\mathbf{p}_{j}, \!^{L_{i}}\mathbf{n}_{j}\! \right) \!\!=\!\! {{^\mathcal{M}\mathbf{u}_{j}}\!^{T}}\!\! \left(\!{^\mathcal{M}\mathbf{T}_{I_{i}}} \!{^{I_{i}}\mathbf{T}_{L_{i}}} \!\! \left(\!^{L_{i}}\mathbf{p}_{j}\!+\! ^{L_{i}}\mathbf{n}_{j} \!\right) \!\!-\!\! ^\mathcal{M}\mathbf{q}_{j}\!\right)  \eqno{(6)}
$$
where \( ^{L_{i}}\mathbf{p}_{j} \) and \( ^{L_{i}}{\mathbf n}_j \) are the LiDAR point $j$ in current $i$-th frame of point cloud and its noise, and we assume that this noise is affected by zero-mean Gaussian white noise. \( ^\mathcal{M}\mathbf{u}_j \) is the normal vector of the associated plane fitted using neighboring points of \( ^{L_{i}}\mathbf{p}_{j} \) in the map \( \mathcal{M} \). \( ^\mathcal{M}\mathbf{T}_{I_{i}} \) is the transformation between the map frame \( \mathcal{M} \) and the IMU frame \( \mathbf{I}_{i} \). \( ^{I_{i}}\mathbf{T}_{L_{i}} \) is the transformation between the IMU frame \( \mathbf{I}_{i} \) and the LiDAR frame \( \mathbf{L}_{i} \). \( ^\mathcal{M}\mathbf{q}_{j} \) is another point on the associated fitted plane in the map \( \mathcal{M} \).

Then, a dual-map based localization scheme tailored for complex, expansive and dynamic environments is introduced. Unlike methods that solely rely on prior maps for pose estimation and correction, this approach also employs a tightly coupled LiDAR-inertial odometry integrated with a local online dynamic map. Meanwhile, instead of running a separate odometry system, we achieve tightly coupled feature-to-map matching by associating point cloud features with both the static prior map \( \mathcal{M}_{global} \) and the real-time local dynamic map \( \mathcal{M}_{local} \) within a single frame processing. 
$$
\mathbf{0}\!=\!^{\mathcal{M}_{global}}\mathbf{R}_{j} \! \left(\!\mathbf{x}_{i}, \!^{L_{i}}\mathbf{p}_{j}, \!^{L_{i}}\mathbf{n}_{j}\!\right)\! + \!^{\mathcal{M}_{local}}\mathbf{R}_{j} \!\left(\!\mathbf{x}_{i}, \!^{L_{i}}\mathbf{p}_{j}, \!^{L_{i}}\mathbf{n}_{j}\!\right)  \eqno{(7)}
$$
This integration enables more precise, reliable, and robust localization. 

Besides, to further accelerate computation, the new mathematically equivalent formula for calculating the Kalman gain is used, as proposed by FAST-LIO \cite{9372856}. This reduces the computational complexity from the observation dimension to the state dimension, ensuring that the measurement update process remains computationally efficient and effective despite the large size of LiDAR point cloud observation dimension.

\subsection{IMU Linear Acceleration and Angular Velocity Updates}
\label{sec:imu_correction}

In addition to performing the observation update using LiDAR data, the idea from Point-LIO \cite{Point-LIO} is incorporated, which treats IMU data as a measurement for filter correction. Specifically, assuming that the IMU measurement is affected by zero-mean Gaussian white noise, and considering that IMU's linear acceleration, angular velocity and their biases are included in the system state estimation, a measurement update can be performed upon receiving IMU data:
\begin{align*}
\boldsymbol{\omega} &= ^{I}\!\!\boldsymbol{\omega} - \mathbf{n}_{\boldsymbol{\omega}} - \mathbf{b}_{\boldsymbol{\omega}} - \mathbf{n}_{\mathbf{b}\boldsymbol{\omega}}  \tag{8} \\
\mathbf{a} &= ^{I}\!\!\mathbf{a} - \mathbf{n}_{\mathbf{a}} - \mathbf{b}_{\mathbf{a}} - \mathbf{n}_{\mathbf{ba}}  \tag{9}
\end{align*}
where \( ^{I}\boldsymbol{\omega} \) and \( ^{I}\mathbf{a} \) are the measurements of IMU angular velocity and linear acceleration. 

Then, for the $i$-th frame of IMU measurement data, we can denote the IMU constraints as:
\begin{align*}
&\mathbf{R}\left(\mathbf{x}_{i}, ^{I}\!\!\boldsymbol{\omega}_{i}, \mathbf{n}_{\boldsymbol{\omega}} \right) = ^{I}\!\!\boldsymbol{\omega}_{i} - \mathbf{n}_{\boldsymbol{\omega}} - \mathbf{b}_{\boldsymbol{\omega}} - \mathbf{n}_{\mathbf{b}\boldsymbol{\omega}} - \boldsymbol{\omega}_{i}   \tag{10} \\
&\mathbf{R}\left(\mathbf{x}_{i}, ^{I}\!\!\mathbf{a}_{i}, \mathbf{n}_{\mathbf{a}} \right) \,\,= ^{I}\!\!\mathbf{a}_{i} - \mathbf{n}_{\mathbf{a}} - \mathbf{b}_{\mathbf{a}} - \mathbf{n}_{\mathbf{ba}} - \mathbf{a}_{i}  
  \tag{11}
\end{align*}

These IMU constraints can be integrated in the following form for the observation update:
$$
\mathbf{0} = \mathbf{R}\left(\mathbf{x}_{i}, \boldsymbol{\omega}_{i}^{I}, \mathbf{n}_{\boldsymbol{\omega}} \right) + \mathbf{R}\left(\mathbf{x}_{i}, \mathbf{a}_{i}^{I}, \mathbf{n}_{\mathbf{a}} \right)   \eqno{(12)}
$$

\subsection{Implement Details}
\label{sec:details_imple}

\textbf{LiDAR Motion Compensation}. 
When the LiDAR is in motion, the points are sampled at different poses, which can introduce distortions in the data. To mitigate this, we adopt a motion compensation strategy similar to that used in Fast-LIO \cite{9372856}, leveraging filter state information. This approach involves maintaining a history of filter states over a specified period. For each point in the LiDAR frame, we interpolate this historical data to estimate the exact state at the moment the point was captured. By obtaining this state estimate, we can accurately project each point to a common reference frame, specifically the end of the point cloud scan. This process effectively compensates for distortions caused by the sensor's motion, ensuring that the resulting point cloud is coherent and provides an accurate representation of the environment.

\textbf{Prior Map Construction and Local Map Updates.}
Constructing a LiDAR point cloud map is essential for high-precision localization systems. Initially, point cloud data often contains noise and redundancy. To address this, we use a simple yet effective interval sampling technique for denoising and downsampling, which improves both data quality and processing efficiency.

For the global prior map, we combine the Fast-LIO2 algorithm \cite{9697912} with high-precision inertial navigation system (INS) data to accurately align point clouds from various poses, ensuring the creation of a consistent global map. In large-scale, dynamic environments, the presence of numerous moving objects and structures can cause ghosting and unreliable information. To deal with this issue, we apply post-processing to the global map, removing dynamic elements and only retaining static point cloud data to prevent localization drift.
In the local dynamic map, we integrate the current frame's point cloud data using corrected LiDAR poses. Continuous updates from the iESKF filter help mitigate cumulative localization errors, maintaining the accuracy of the map. Outdated and distant data is also removed to optimize resource usage and ensure efficient output during local point cloud map registration in LiDAR-inertial odometry.

To enhance matching efficiency with both the local dynamic map and the global prior map, we employ the incremental k-d tree (ikd-Tree) data structure \cite{9697912} for incremental map management. The ikd-Tree efficiently handles point insertion, deletion, dynamic rebalancing, and nearest neighbor searches, significantly improving the speed and accuracy of point cloud registration.

\section{EXPERIMENTS}

To validate the effectiveness of the proposed method, we conduct comprehensive experiments on our private datasets. The experimental setup is detailed in Sec.~\ref{sec:exp_set}, while the localization results are presented in Sec.~\ref{sec:loc_result}.

\subsection{Experiment Setup}
\label{sec:exp_set}

\begin{figure}[htbp]
    \centering
    \begin{subfigure}[b]{0.20\textwidth} 
        \centering 
        \includegraphics[height=3cm,keepaspectratio]{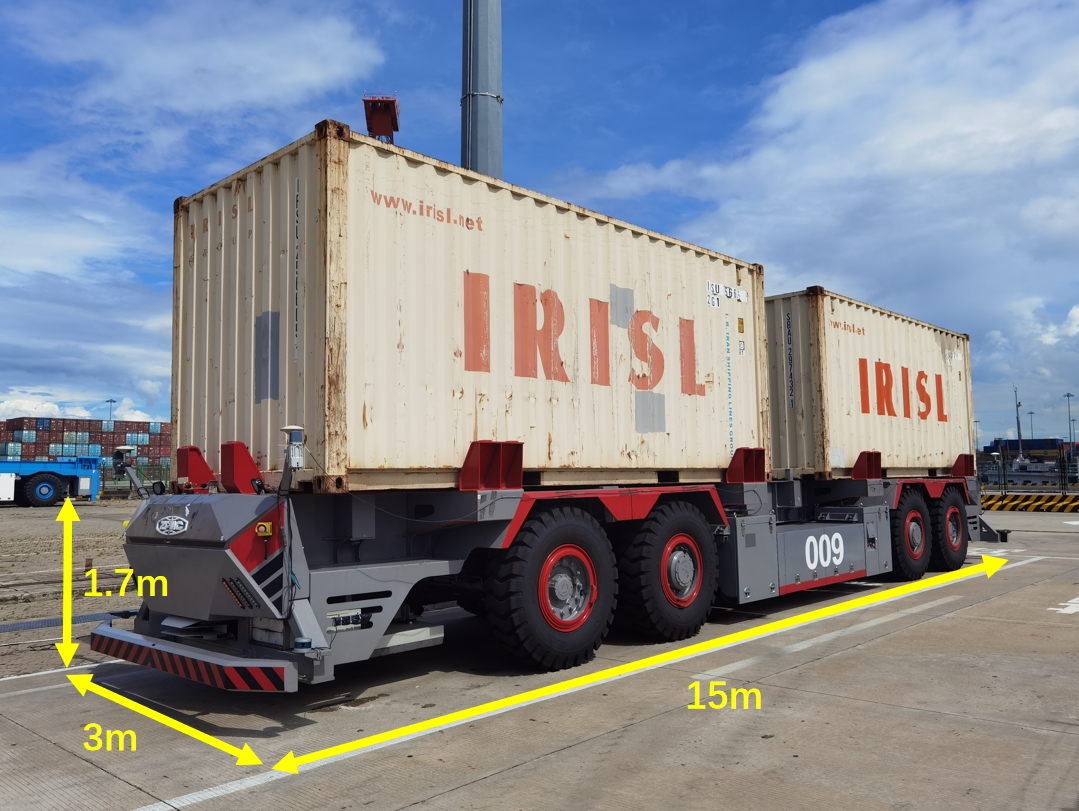} 
        \caption{} 
        \label{fig:left} 
    \end{subfigure}
    \hfill
    \begin{subfigure}[b]{0.25\textwidth} 
        \centering
        \includegraphics[width=4cm,height=3cm]{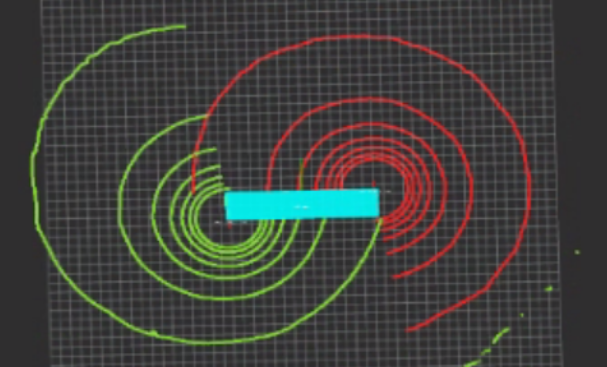} 
        \caption{} 
        \label{fig:right} 
    \end{subfigure}
    \caption{Experimental Setup. (a) The IGV, with a size of 15\,m $\times$ 3\,m $\times$ 1.7\,m. (b) The fused point clouds from two diagonally positioned 16-line LiDARs, where the blue rectangle represents the vehicle body. The green point cloud corresponds to the LiDAR located in the upper-left corner, while the red point cloud originates from the LiDAR in the lower-right corner.}
    \label{fig:side-by-side}
    \vspace{-1 em}
\end{figure}

\begin{figure}[htbp]
    \centering
    \begin{subfigure}[b]{0.235\textwidth} 
        \centering
        \includegraphics[width=\textwidth]{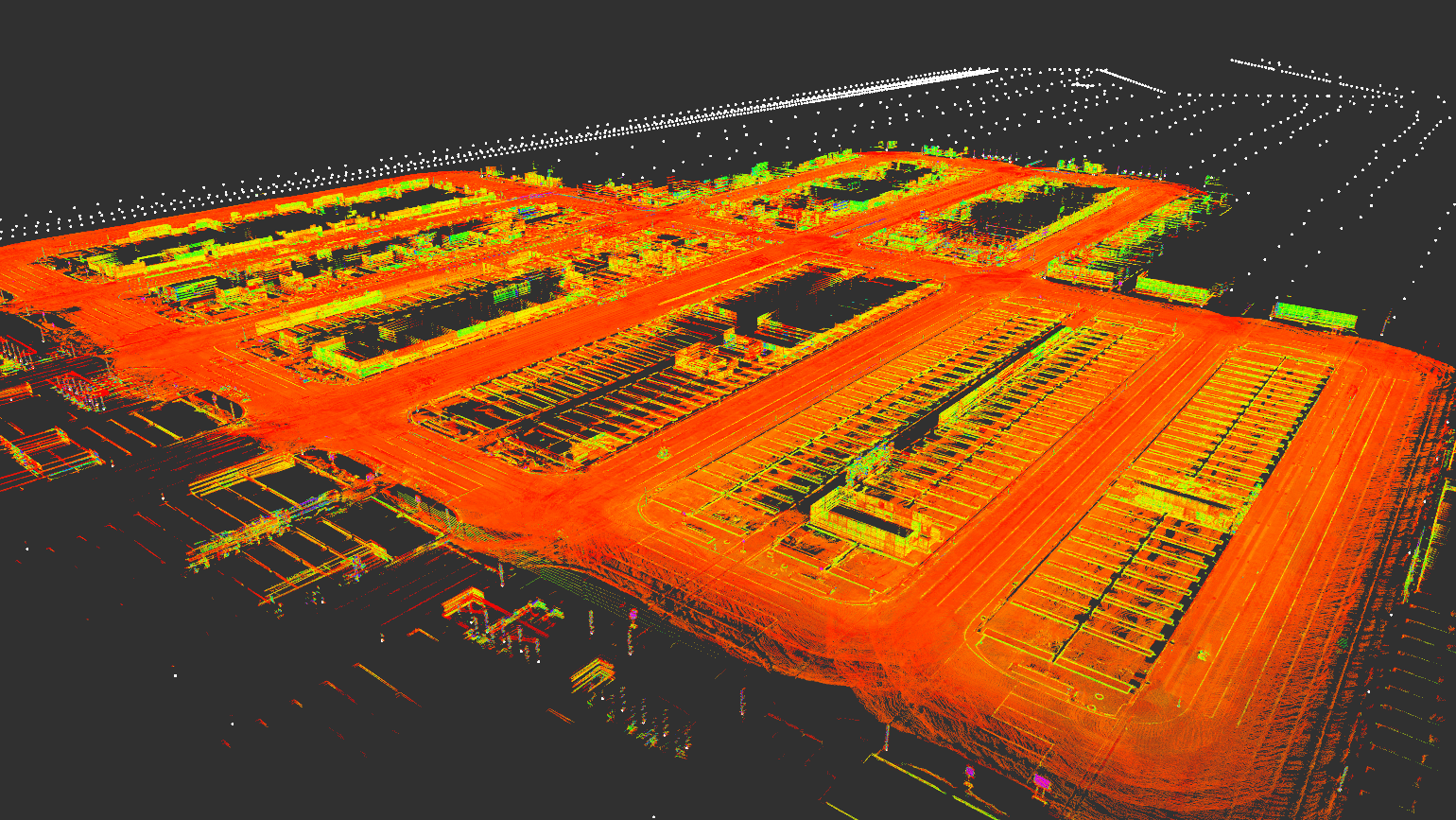} 
        \caption{} 
        \label{fig:top-left} 
    \end{subfigure}
    \hfill 
    \begin{subfigure}[b]{0.235\textwidth} 
        \centering
        \includegraphics[width=\textwidth]{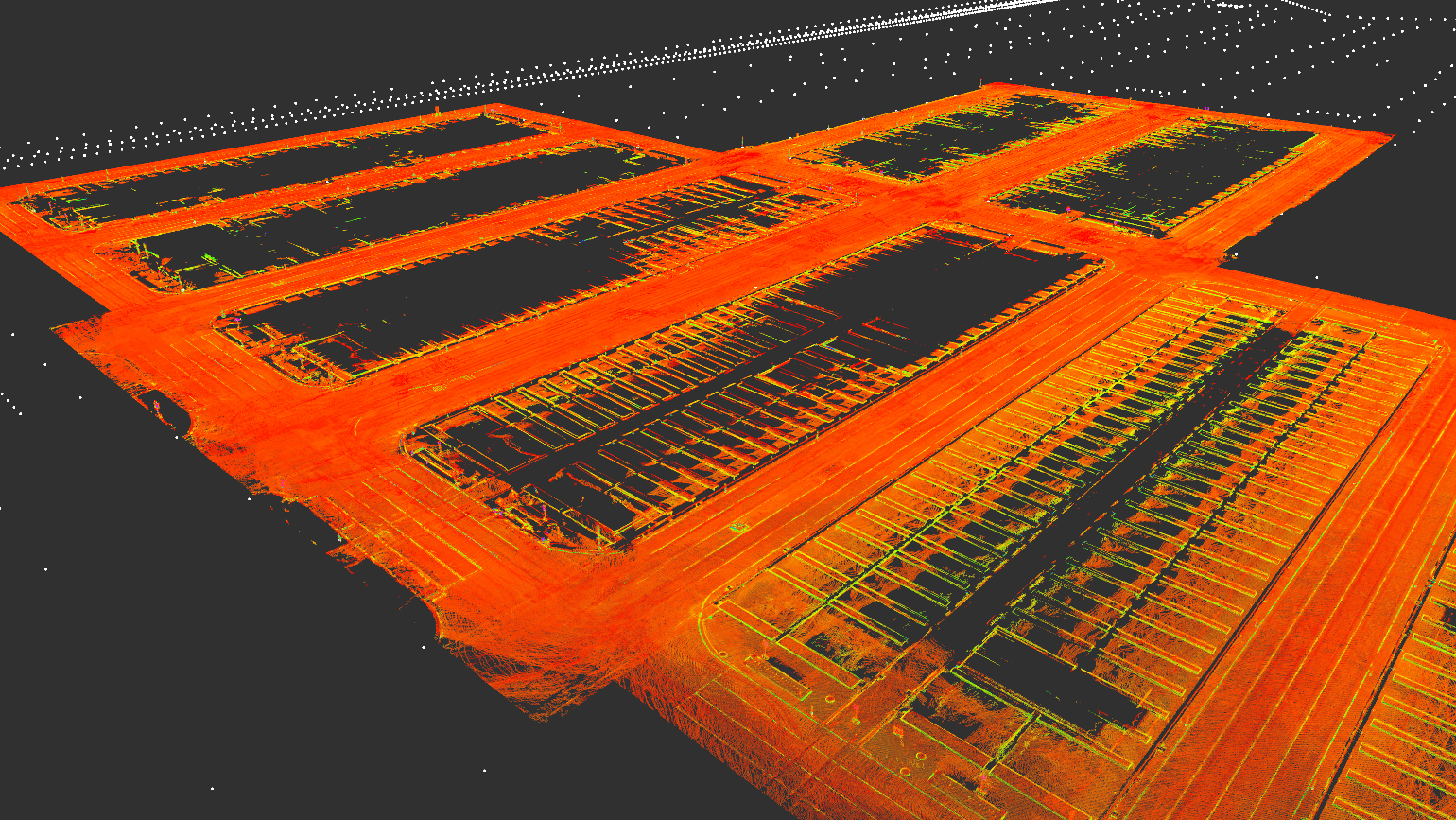} 
        \caption{} 
        \label{fig:top-right} 
    \end{subfigure}
    
    \vspace{0cm} 
    \begin{subfigure}[b]{0.455\textwidth} 
        \centering
        \includegraphics[width=\textwidth]{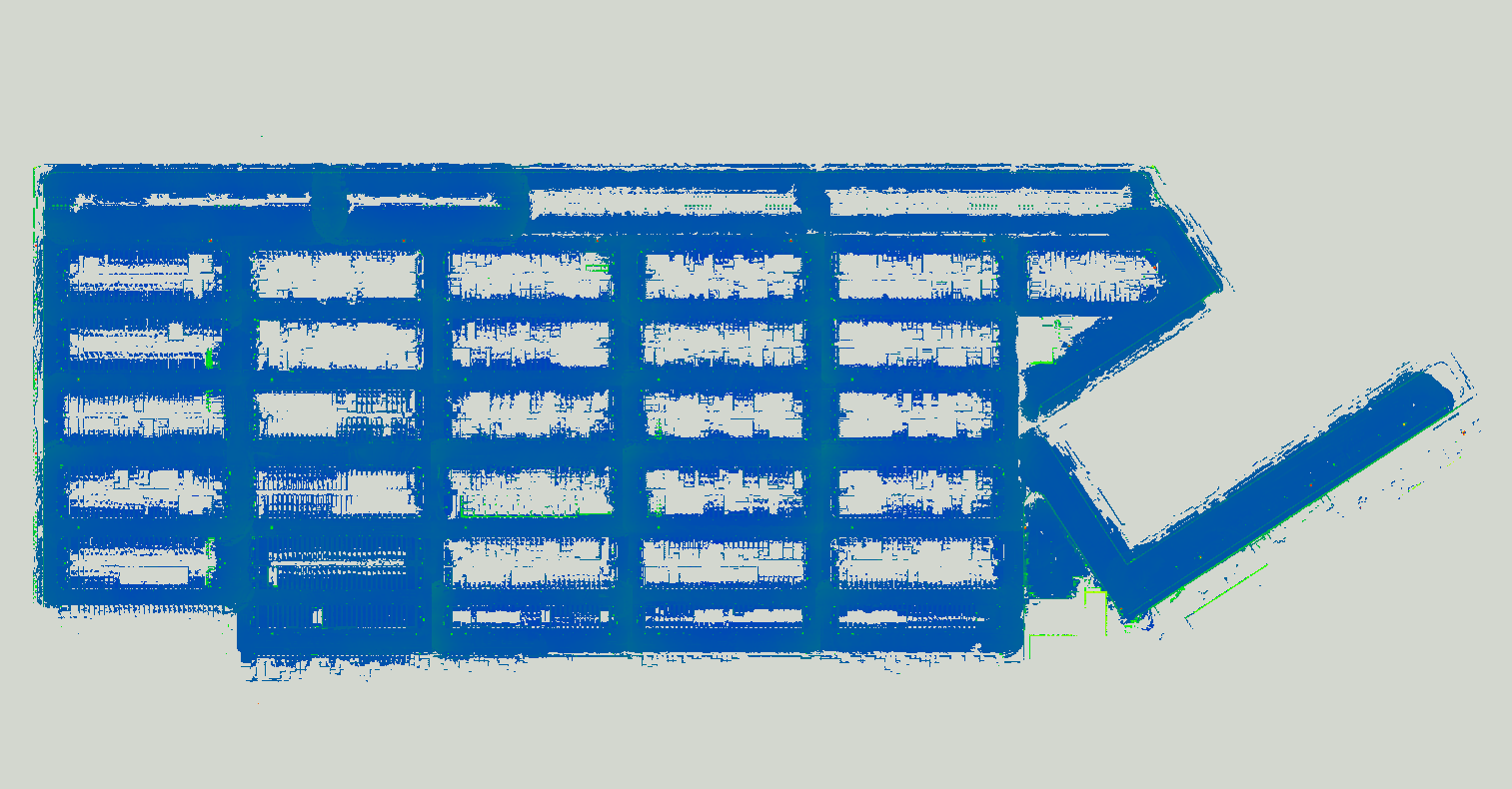} 
        \caption{} 
        \label{fig:bottom} 
    \end{subfigure}
    
    \caption{Point cloud map. (a) Example of the point cloud map without dynamic obstacle removal. (b) Example of the point cloud map with dynamic obstacle removal. (c) The resulting point cloud output with a dimension of 1538\,m $\times$ 596\,m.}
    \label{fig:three-images}
    \vspace{-1 em}
\end{figure}

\begin{figure*}[htbp]
    \centering
    \begin{subfigure}{0.325\textwidth}
        \includegraphics[width=\linewidth,height=0.5\linewidth]{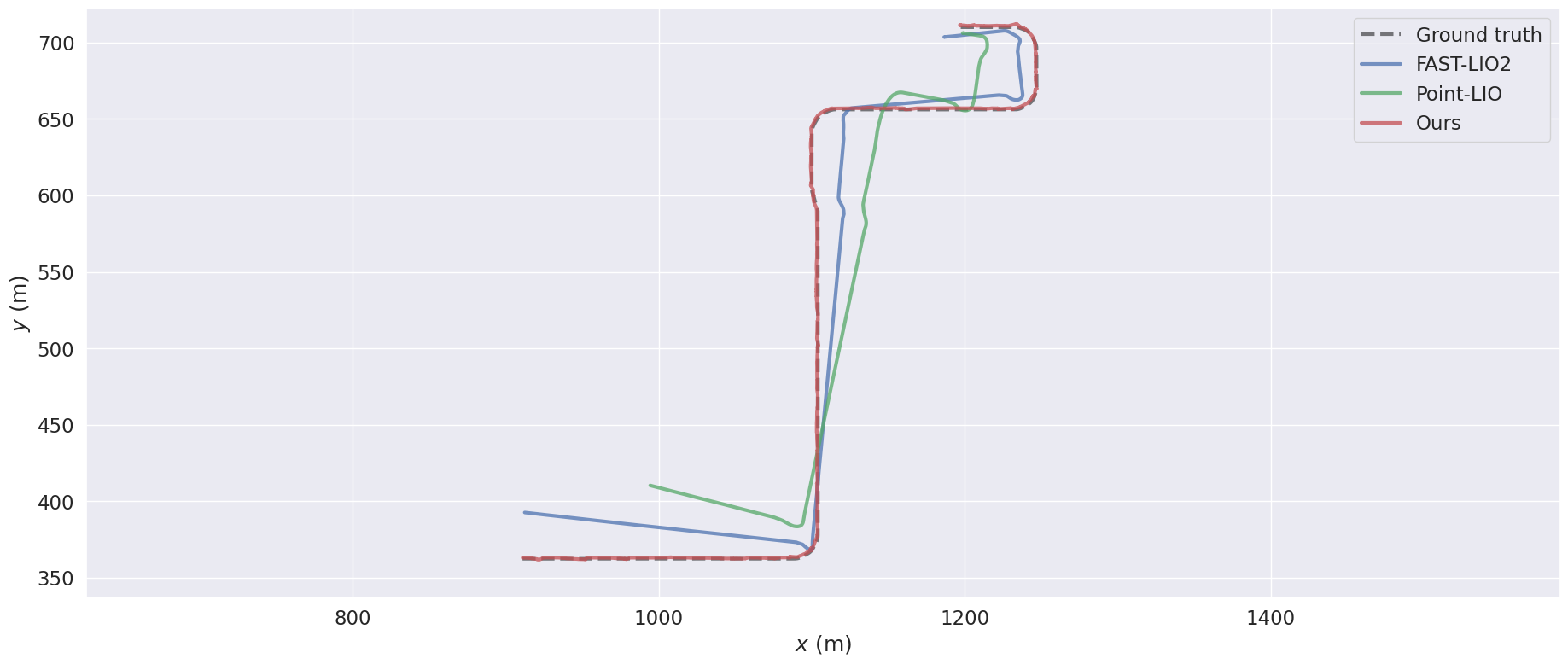}
        \caption{}
    \end{subfigure}
    \hfill
    \begin{subfigure}{0.325\textwidth}
        \includegraphics[width=\linewidth,height=0.5\linewidth]{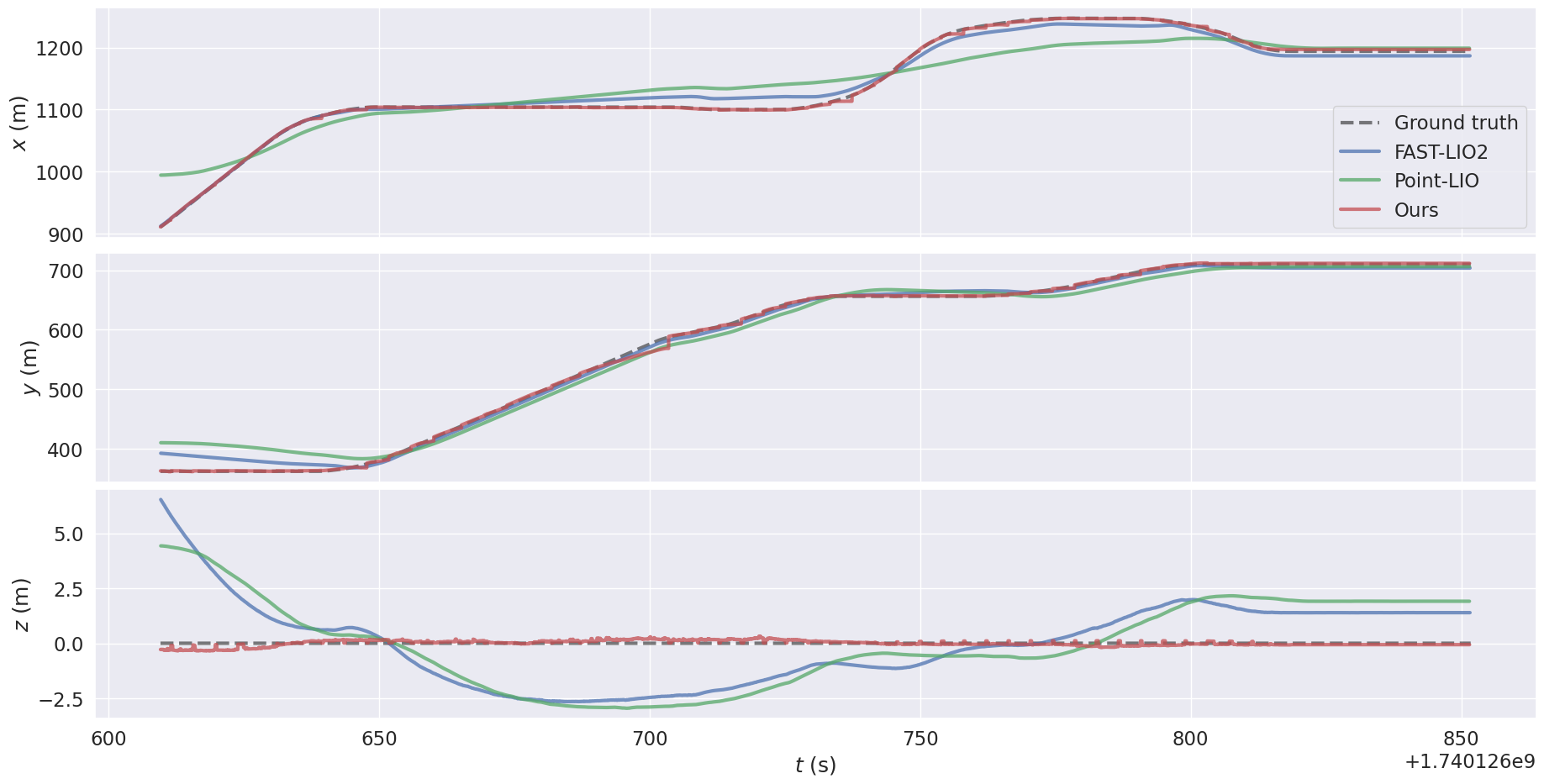}
        \caption{}
    \end{subfigure}
    \hfill
    \begin{subfigure}{0.325\textwidth}
        \includegraphics[width=\linewidth,height=0.5\linewidth]{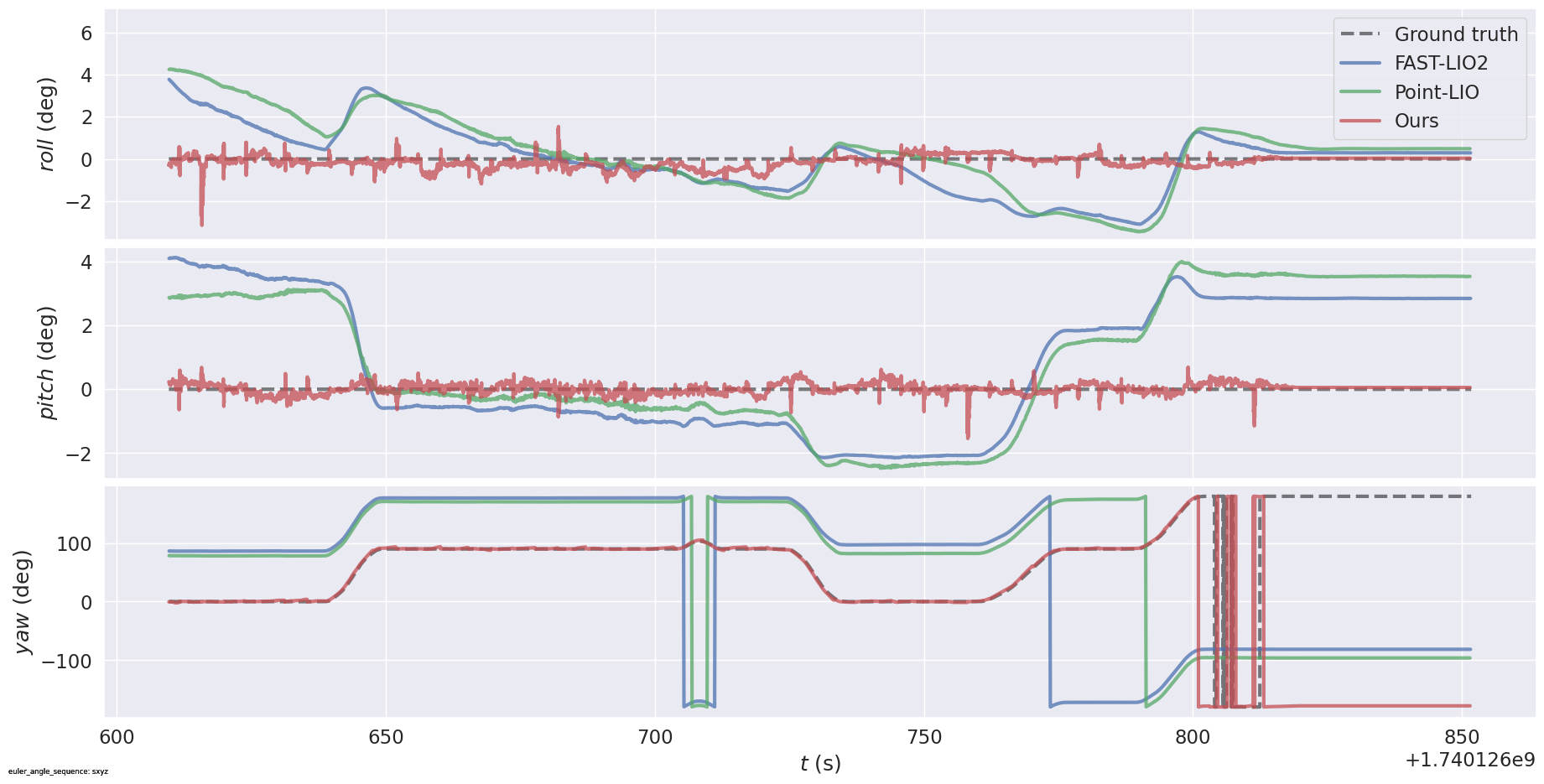}
        \caption{}
    \end{subfigure}

    \begin{subfigure}{0.325\textwidth}
        \includegraphics[width=\linewidth,height=0.5\linewidth]{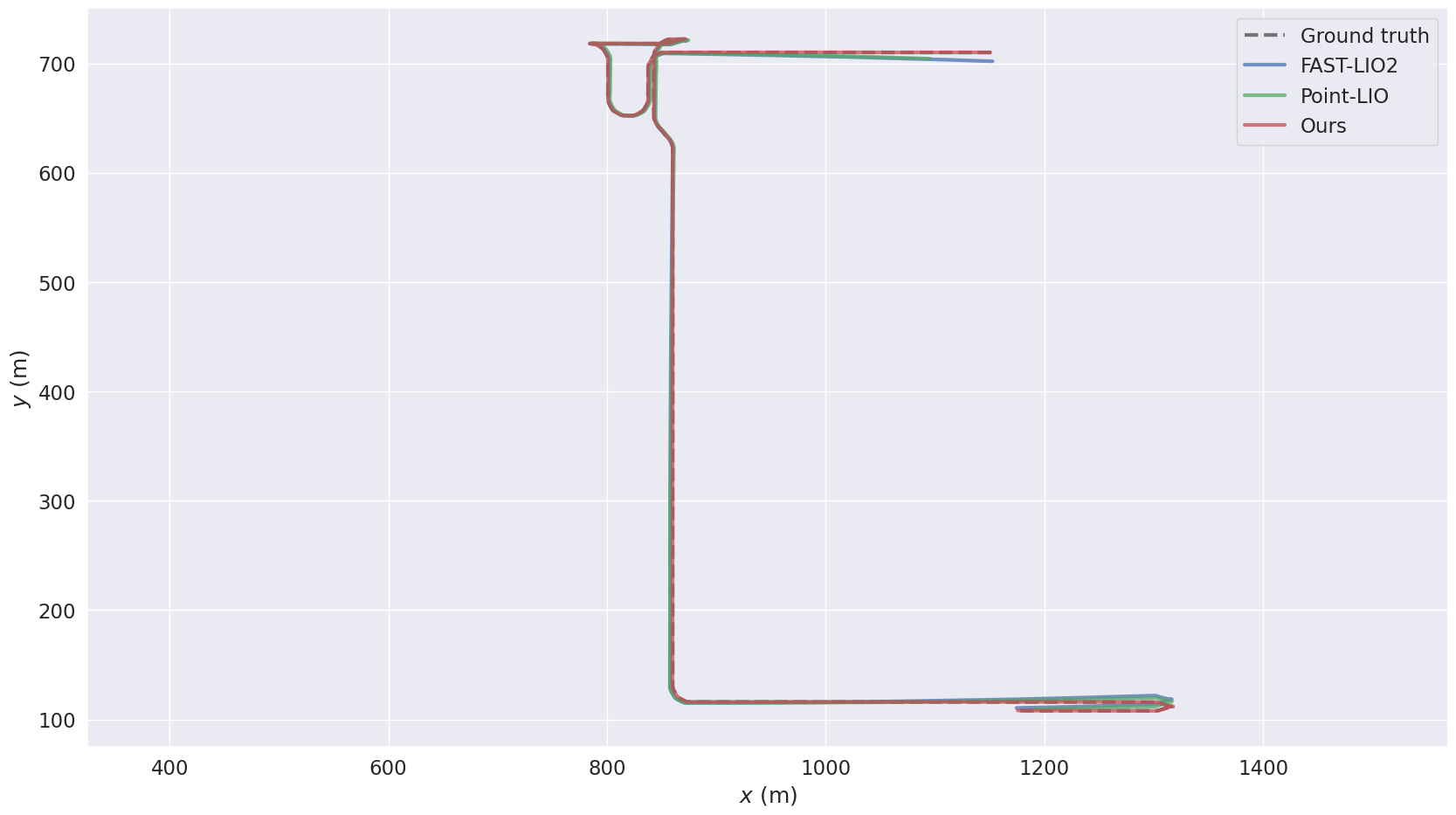}
        \caption{}
    \end{subfigure}
    \hfill
    \begin{subfigure}{0.325\textwidth}
        \includegraphics[width=\linewidth,height=0.5\linewidth]{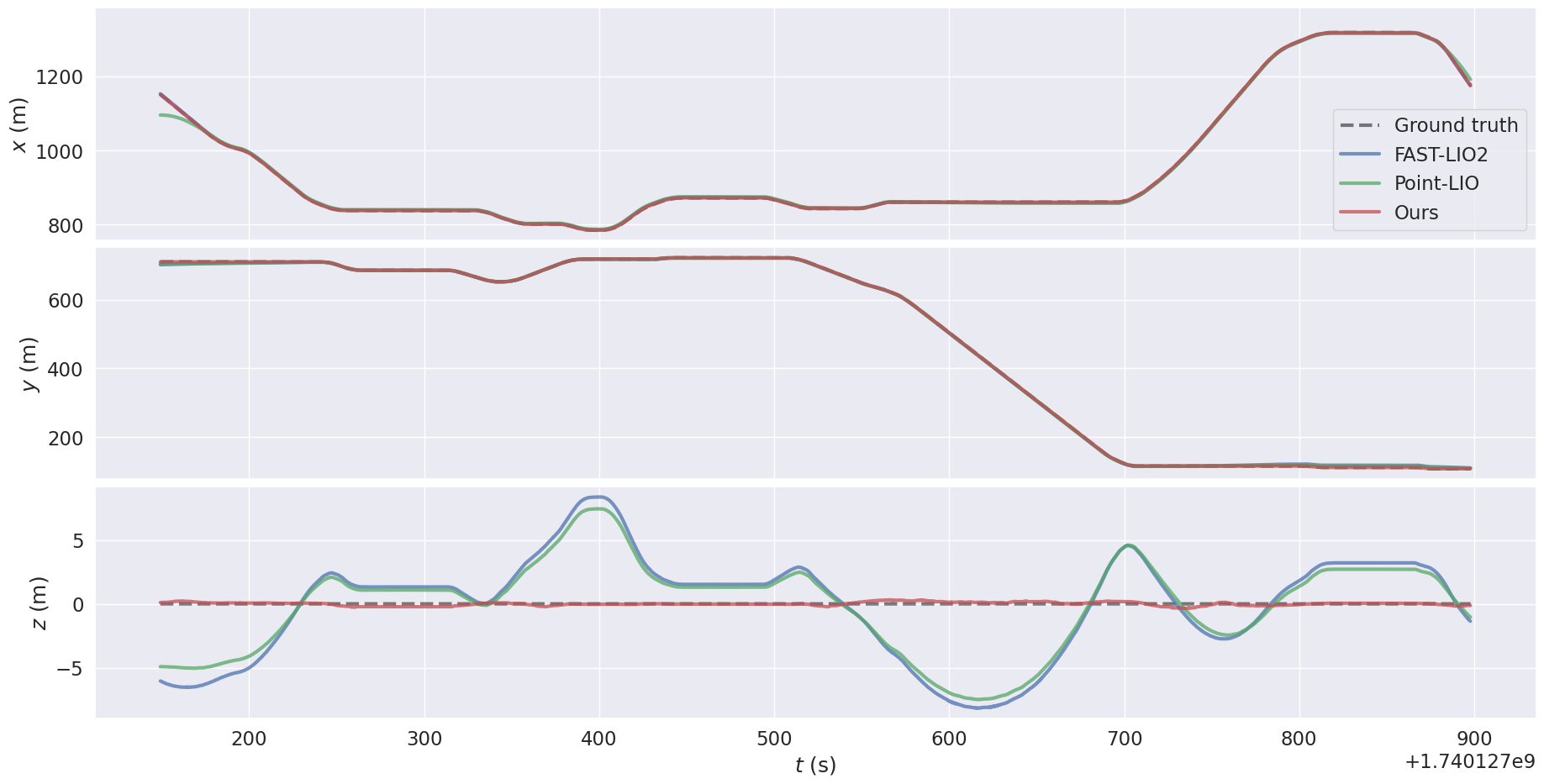}
        \caption{}
    \end{subfigure}
    \hfill
    \begin{subfigure}{0.325\textwidth}
        \includegraphics[width=\linewidth,height=0.5\linewidth]{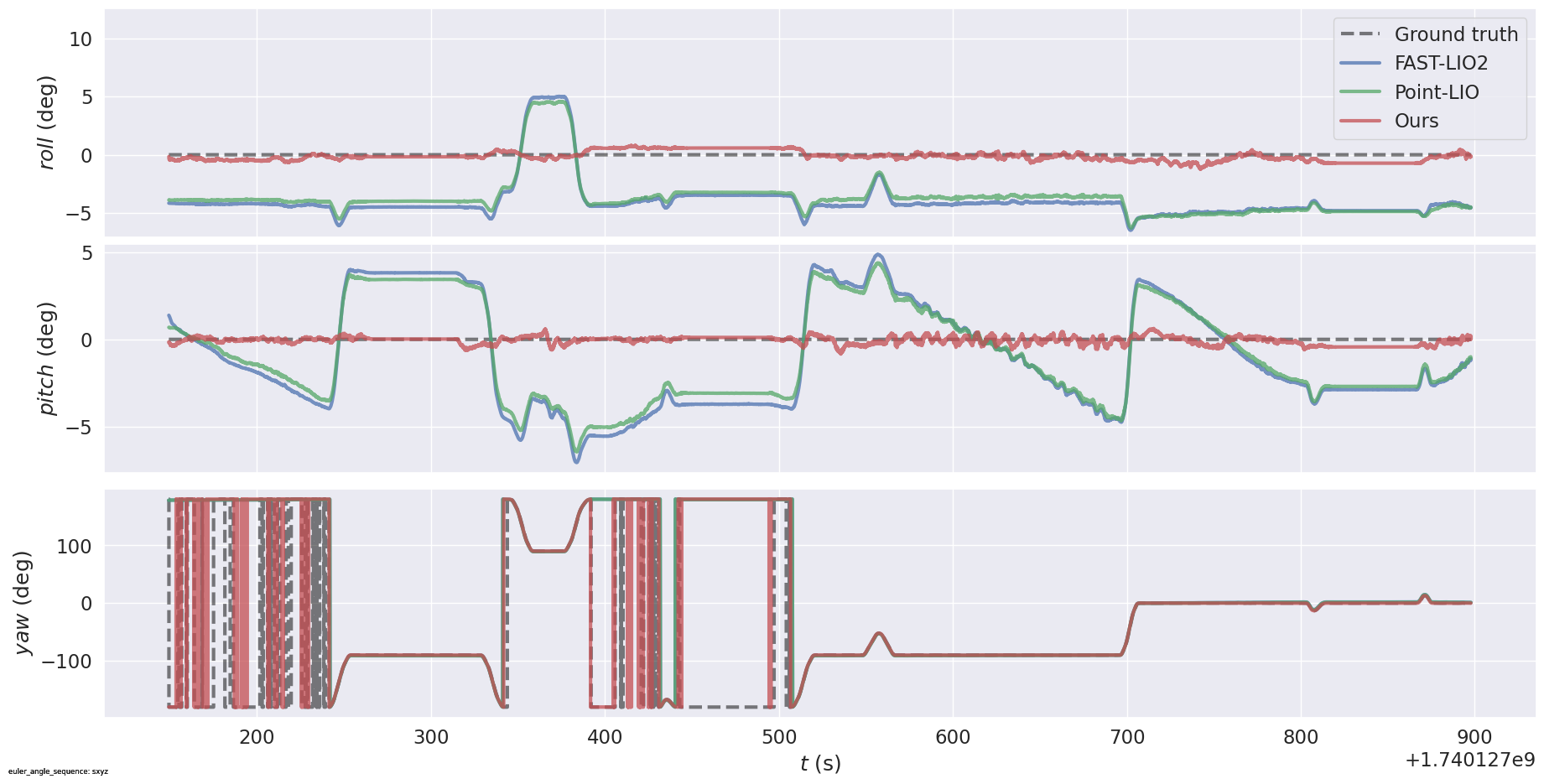}
        \caption{}
    \end{subfigure}

    \begin{subfigure}{0.325\textwidth}
        \includegraphics[width=\linewidth,height=0.5\linewidth]{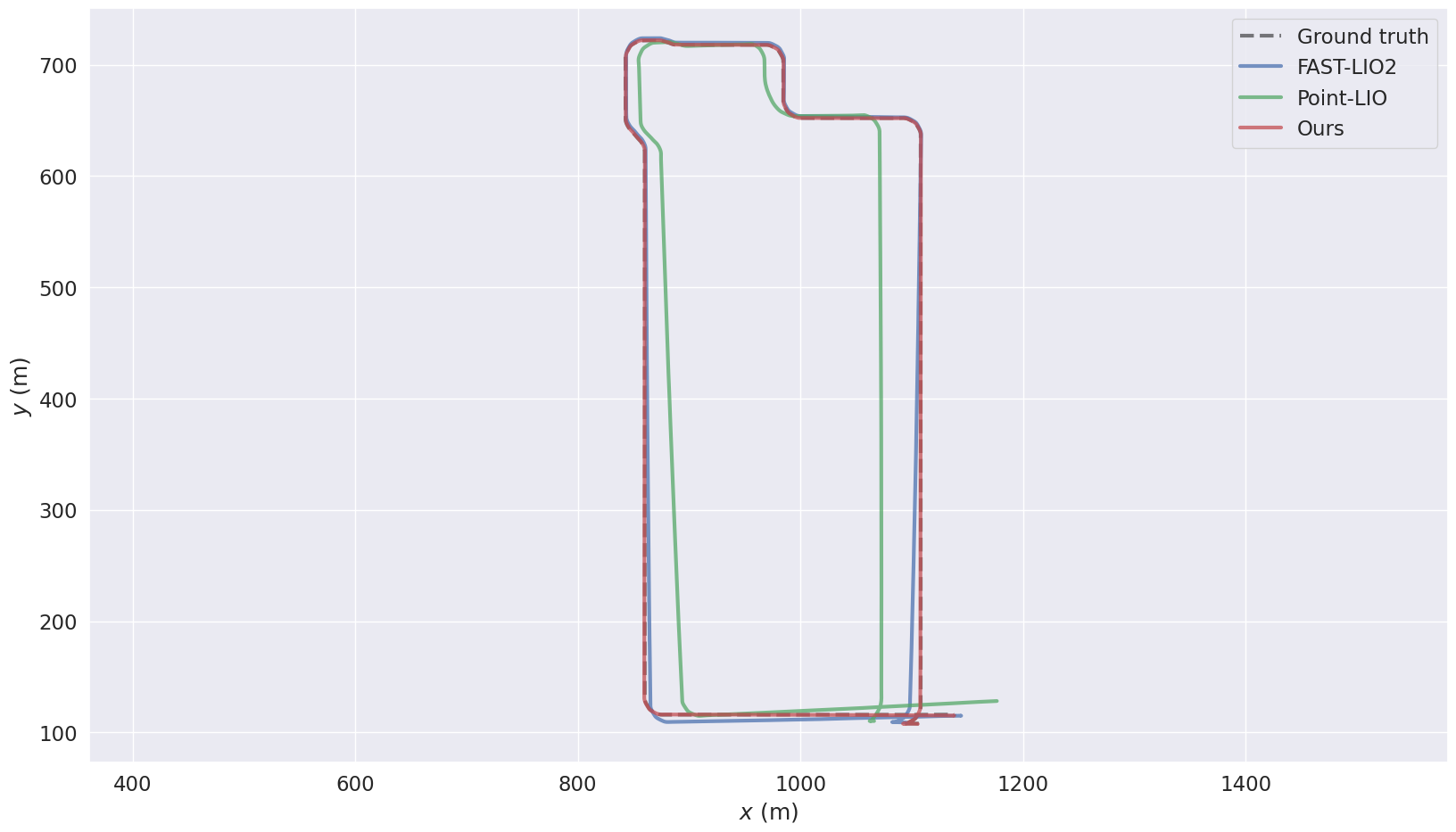}
        \caption{}
    \end{subfigure}
    \hfill
    \begin{subfigure}{0.325\textwidth}
        \includegraphics[width=\linewidth,height=0.5\linewidth]{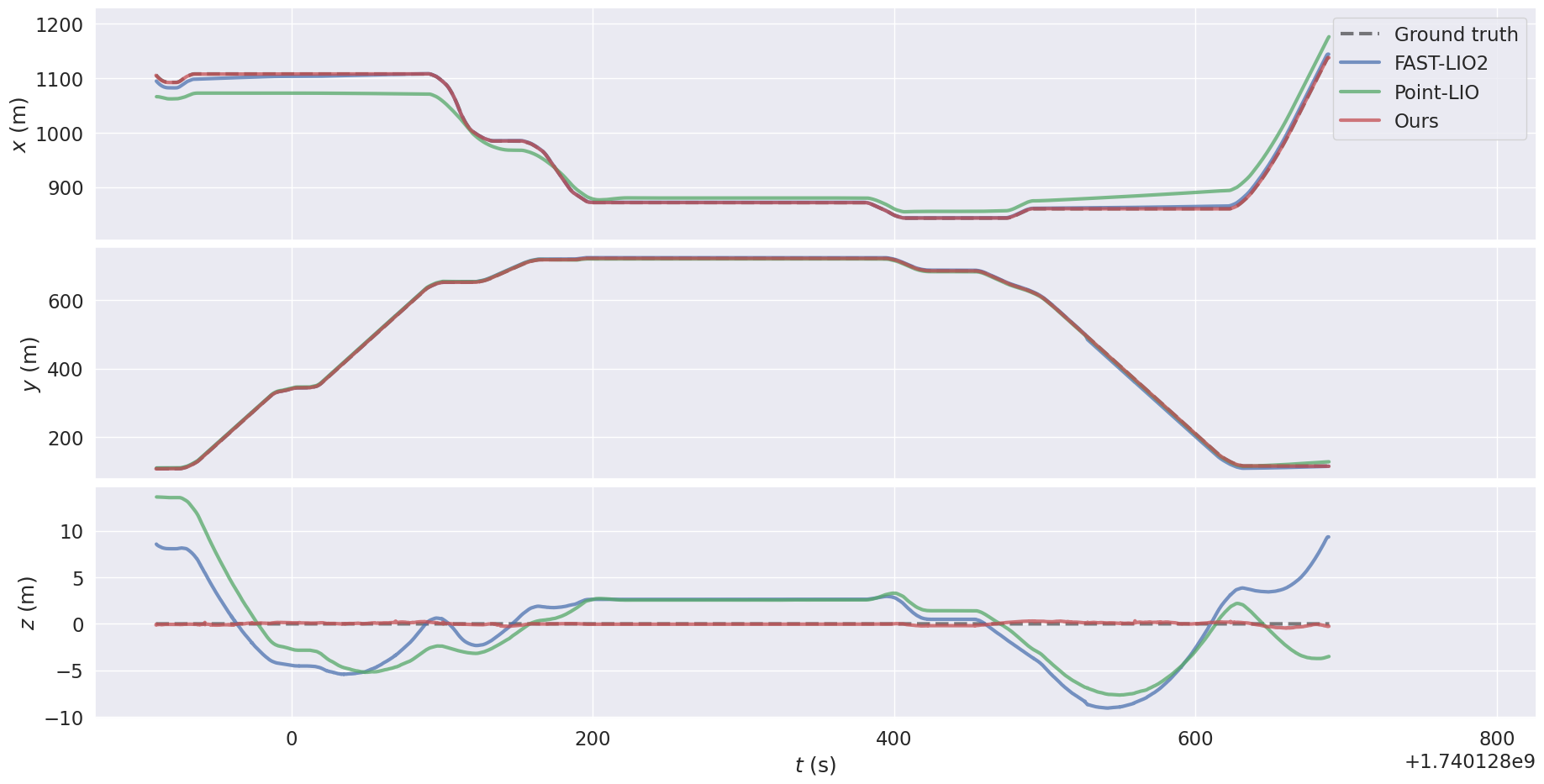}
        \caption{}
    \end{subfigure}
    \hfill
    \begin{subfigure}{0.325\textwidth}
        \includegraphics[width=\linewidth,height=0.5\linewidth]{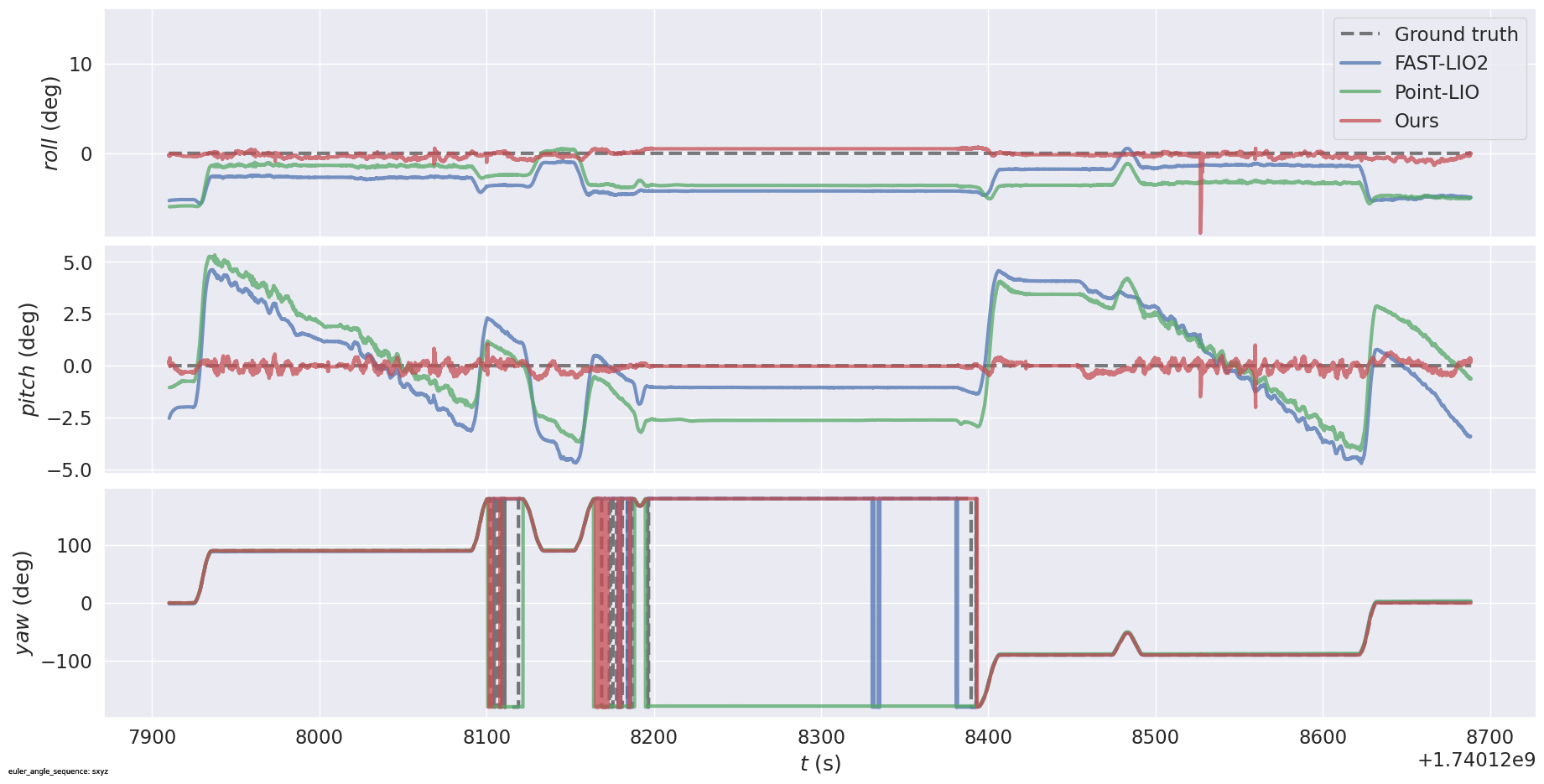}
        \caption{}
    \end{subfigure}

    \begin{subfigure}{0.325\textwidth}
        \includegraphics[width=\linewidth,height=0.5\linewidth]{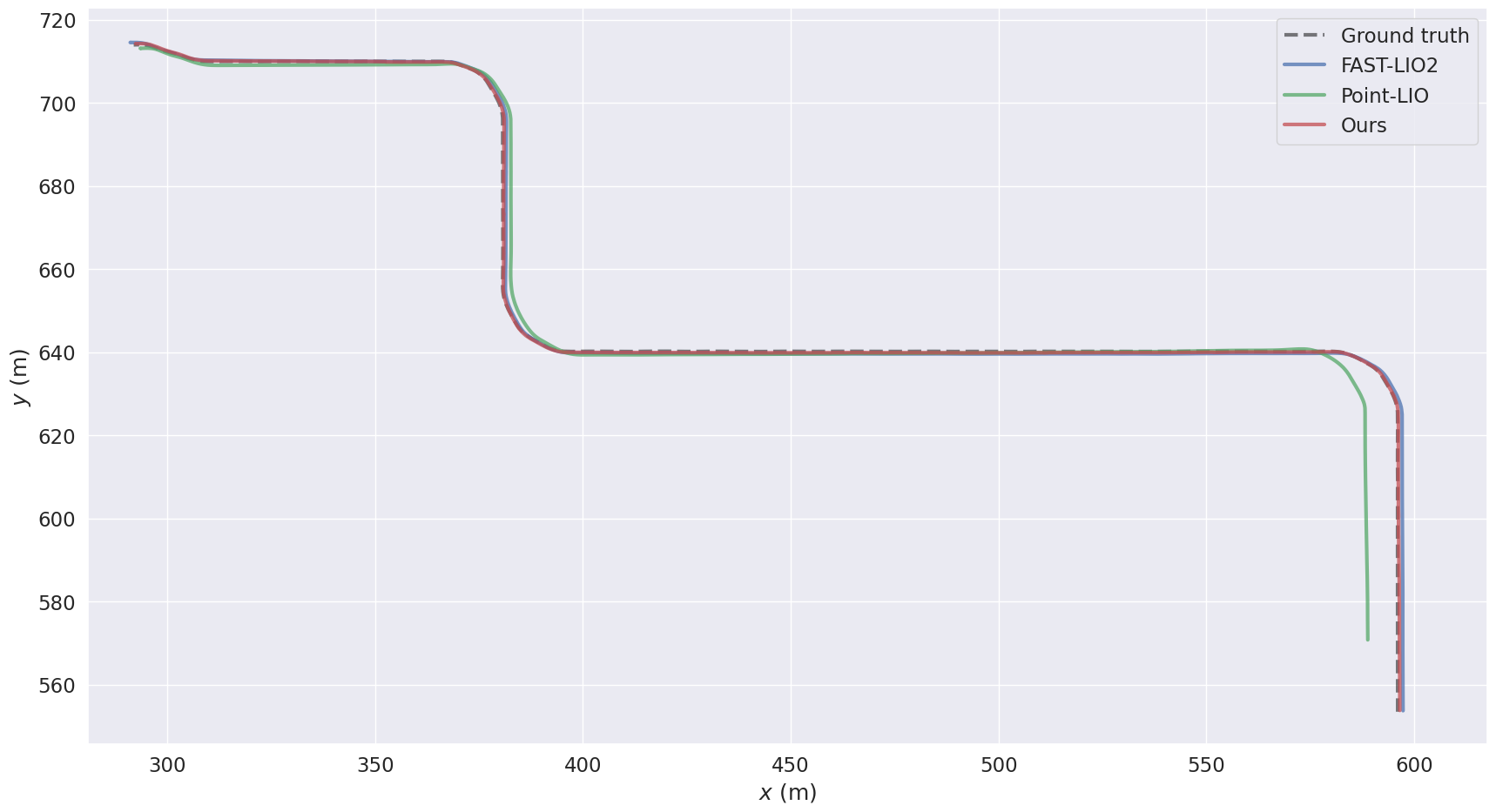}
        \caption{}
    \end{subfigure}
    \hfill
    \begin{subfigure}{0.325\textwidth}
        \includegraphics[width=\linewidth,height=0.5\linewidth]{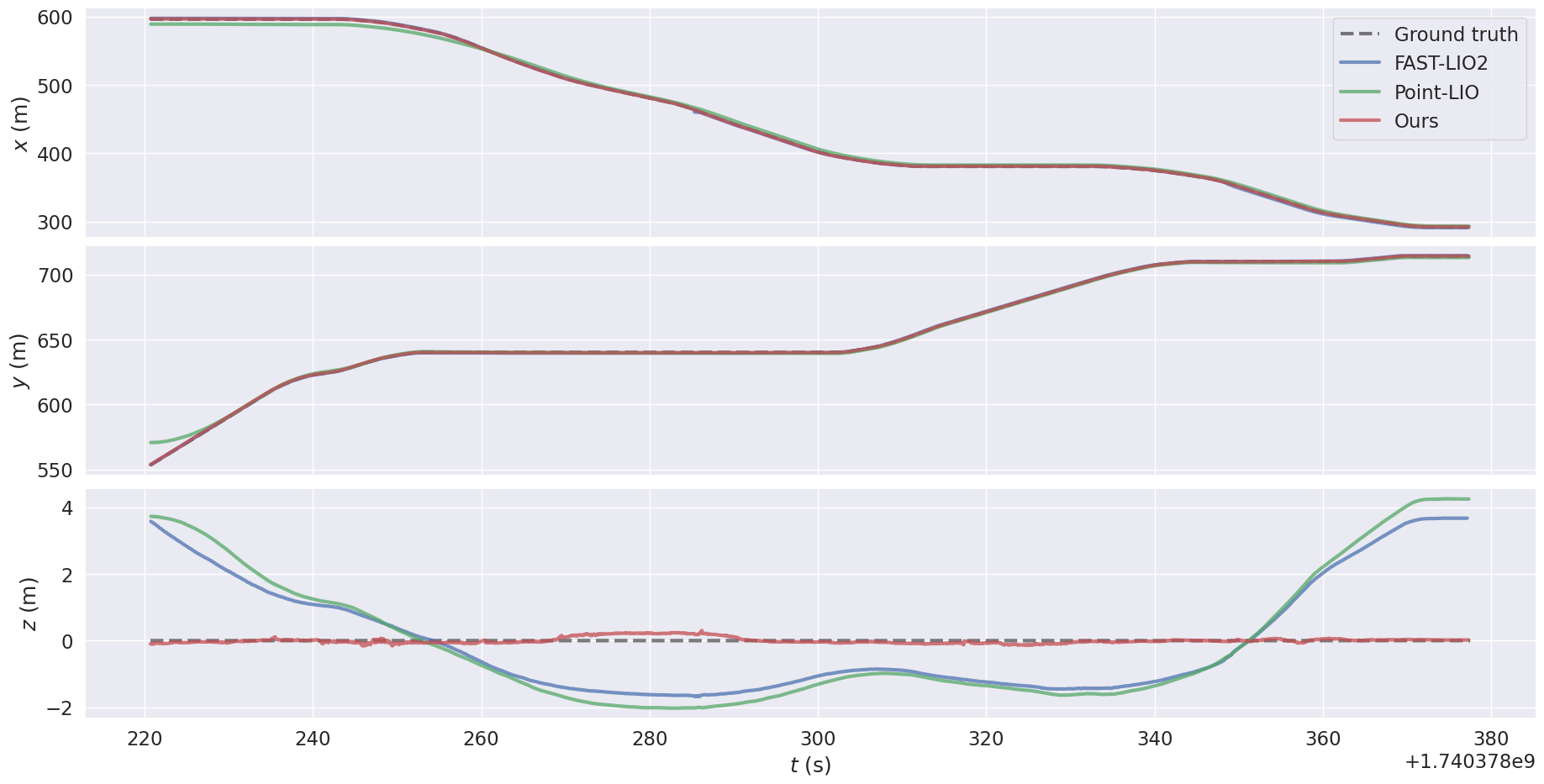}
        \caption{}
    \end{subfigure}
    \hfill
    \begin{subfigure}{0.325\textwidth}
        \includegraphics[width=\linewidth,height=0.5\linewidth]{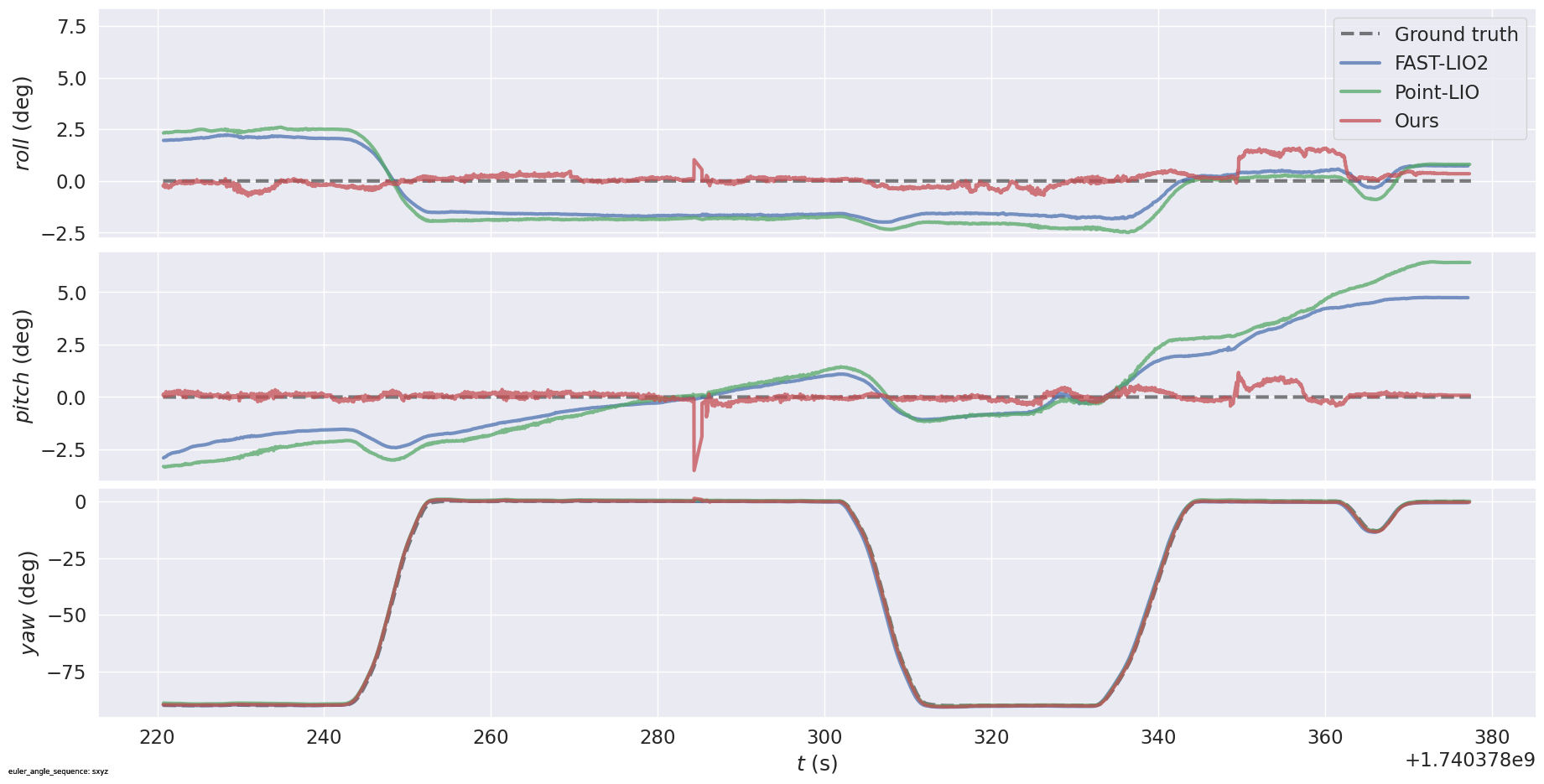}
        \caption{}
    \end{subfigure}

    \begin{subfigure}{0.325\textwidth}
        \includegraphics[width=\linewidth,height=0.5\linewidth]{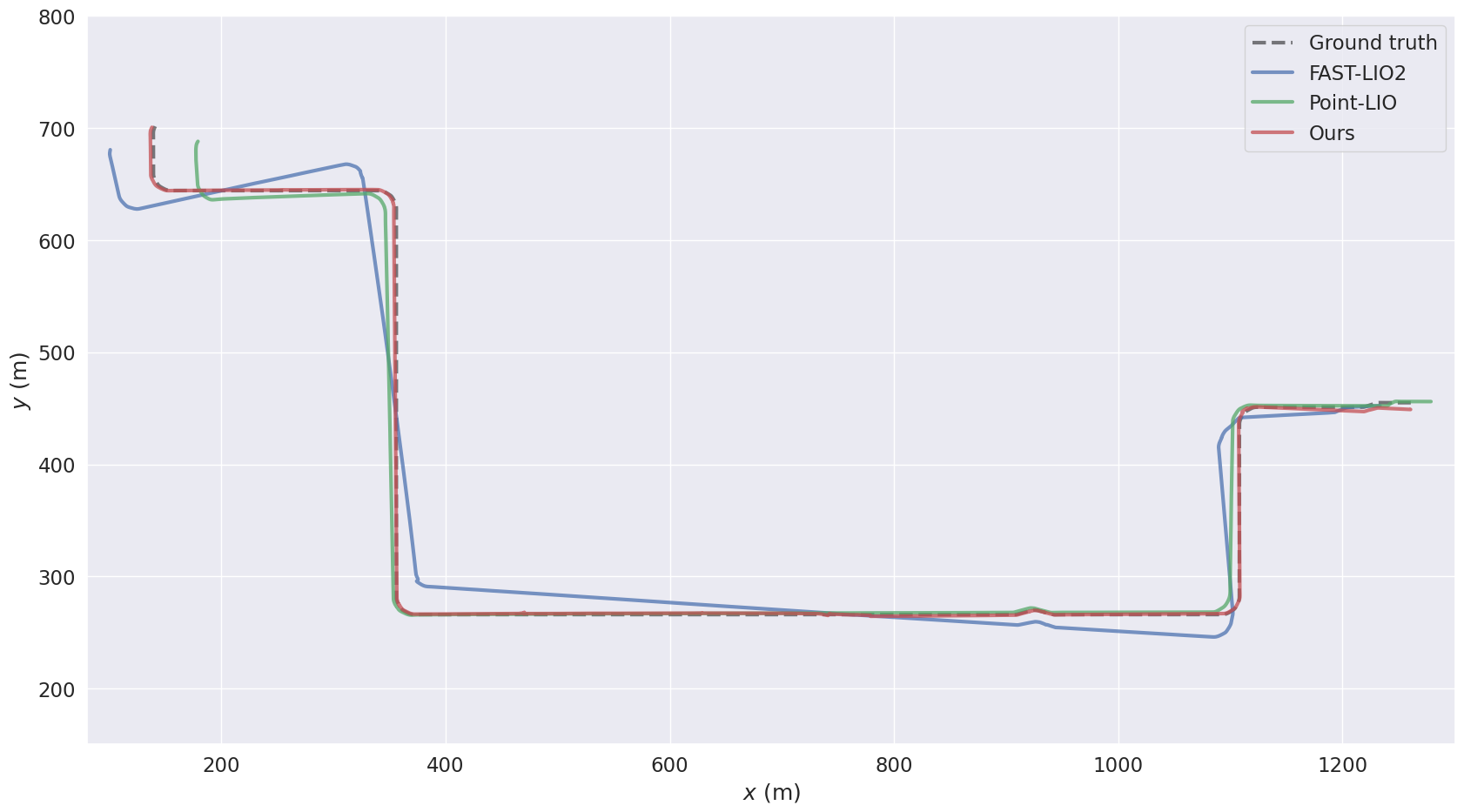}
        \caption{}
    \end{subfigure}
    \hfill
    \begin{subfigure}{0.325\textwidth}
        \includegraphics[width=\linewidth,height=0.5\linewidth]{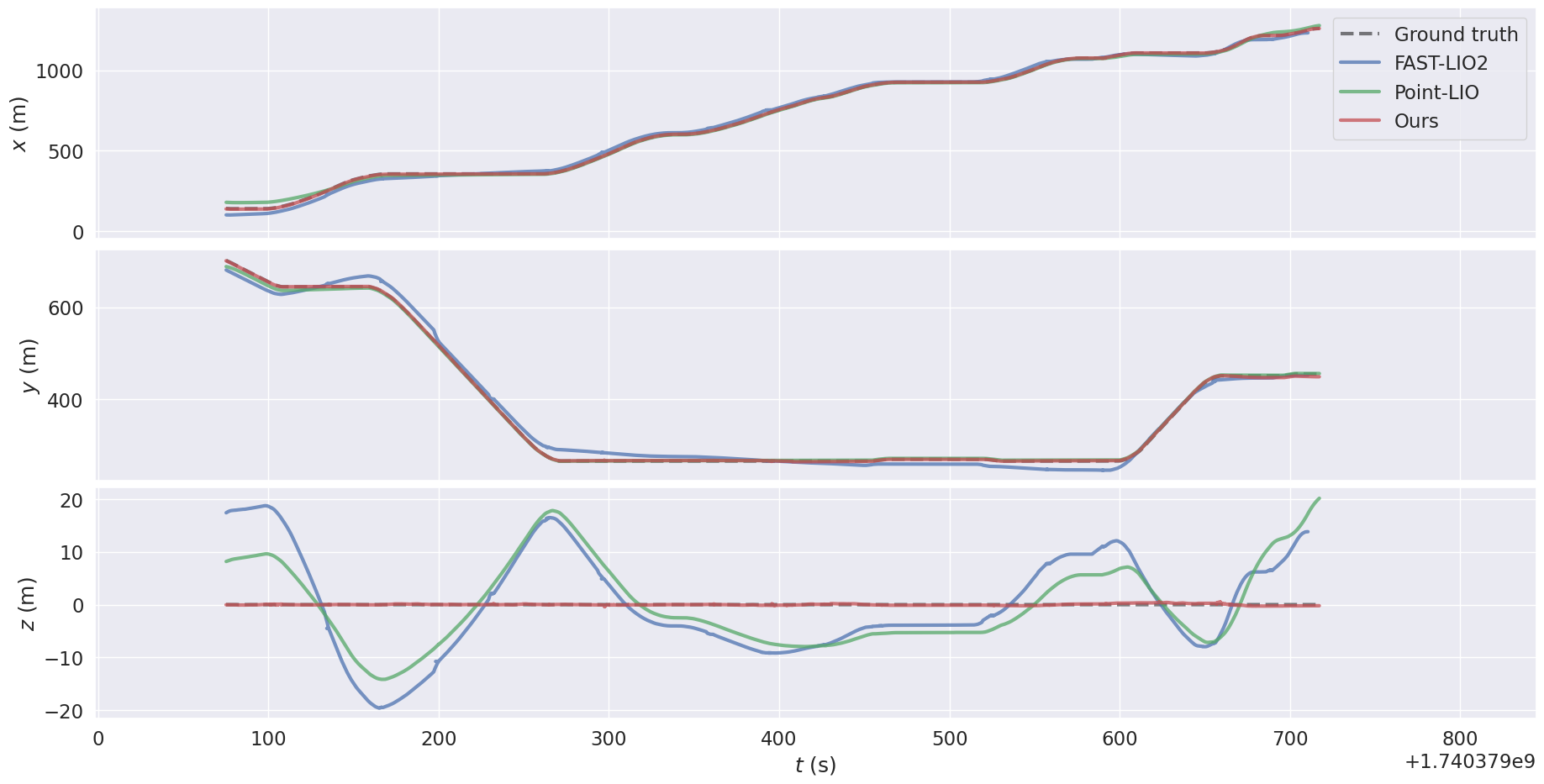}
        \caption{}
    \end{subfigure}
    \hfill
    \begin{subfigure}{0.325\textwidth}
        \includegraphics[width=\linewidth,height=0.5\linewidth]{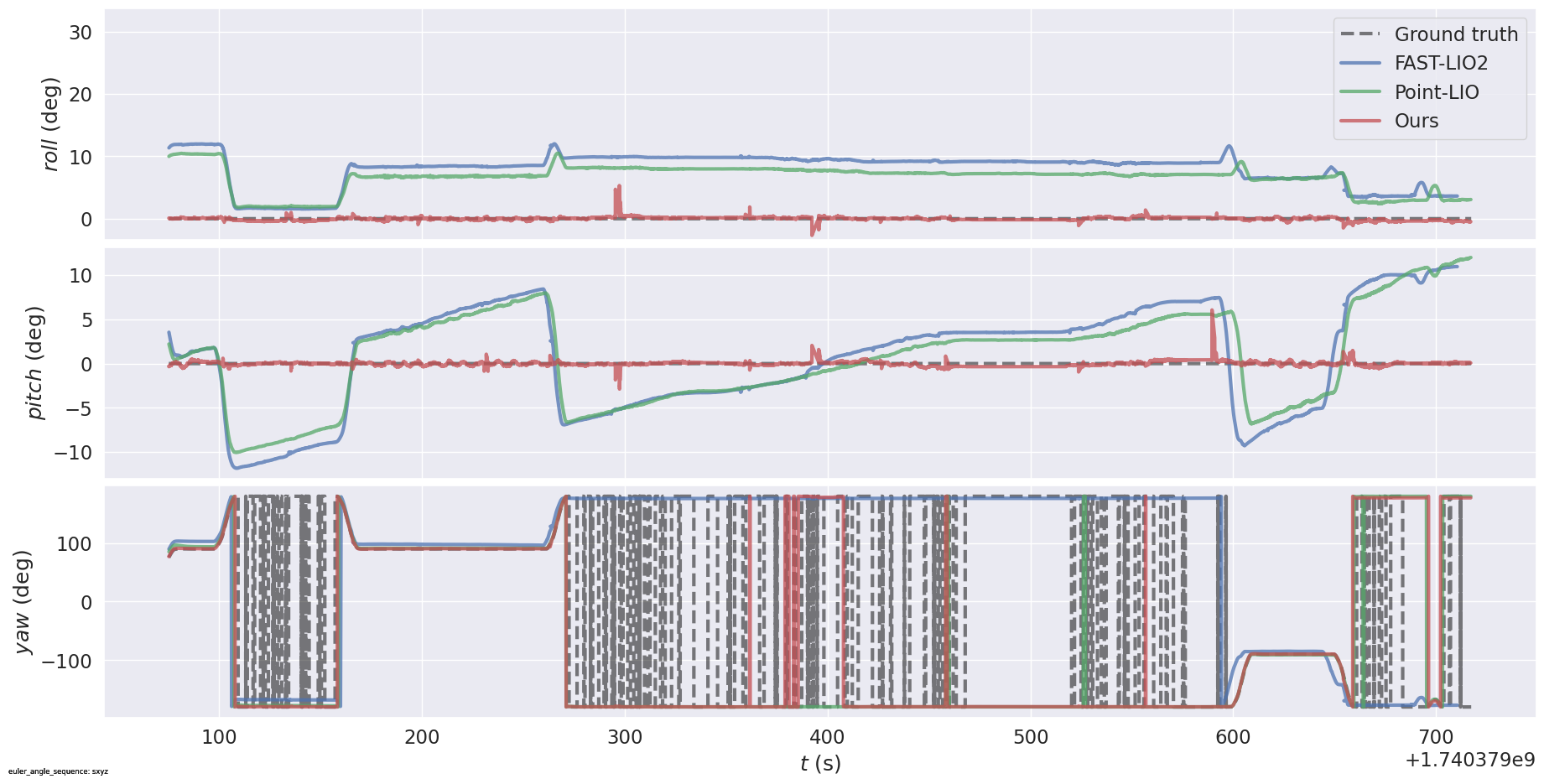}
        \caption{}
    \end{subfigure}

    \caption{Comparison results of the overall trajectories, incorporating both translational and rotational components, across various testing datasets. Each column represents the comparison results of different algorithms on each test dataset. (a), (d), (g), (j), and (m) represent the overall trajectories of each algorithm, (b), (e), (h), (k), and (n) represent the translation components of each algorithm, and (c), (f), (i), (l), and (o) represent the rotation components of each algorithm. Due to the excessive errors in DLL \cite{9636501} and Light-LOAM \cite{10439642}, both methods failed in these scenarios, so we do not present their results.}
    \label{fig:15_images}
    \vspace{-2em}
\end{figure*}

\textbf{Baselines and Metrics.} 
Our baselines include Light-LOAM \cite{10439642}, DLL \cite{9636501}, FAST-LIO2 \cite{9697912}, and Point-LIO \cite{Point-LIO}, which represent state-of-the-art approaches in LiDAR SLAM, localization, and odometry. We perform comparative experiments on these methods to assess the effectiveness of our proposed approach in various scenarios. To ensure a fair comparison, the results of these systems are obtained using the source code provided by their respective authors, with only minor adjustments made to the input data interfaces to adapt to our dataset. To assess the accuracy of SLAM, we use the Absolute Trajectory Error (ATE) of the pose trajectory as the evaluation metric and used the $evo$ toolkit to compare and analyze the localization trajectories of different algorithms. 

\textbf{Datasets and Sensor Equipments}. To evaluate the performance of our method in large-scale, unbounded, and dynamic environments, we conduct a detailed comparison with other state-of-the-art algorithms. We select a port environment as the testing scenario and manually created a private comprehensive dataset. The test area spans approximately 1 million square meters, with the dataset covering a total testing path of 6\,km. Data was collected by driving an IGV throughout the port to capture diverse scenarios and conditions. The dataset includes high-quality 16-line LiDAR and 6-axis IMU data, with the LiDAR operating at a frequency of 10\,Hz and the IMU at 100\,Hz. Additionally, a higher-precision sensor fusion method with RTK, operating at 50\,Hz, is provided as ground truth for quantitatively evaluating the performance of different localization systems. We design experiment trajectories in open areas with few tall mechanical structures to ensure GPS accuracy. To reduce potential LiDAR occlusions, the LiDAR data is generated from the fused point clouds of two 16-line LiDARs, which are time-synchronized and distortion-corrected. These LiDARs are strategically mounted at the upper-left and lower-right corners of the test vehicle to maximize coverage. The test vehicle and the resulting LiDAR point cloud data are illustrated in Fig. \ref{fig:side-by-side}.


\begin{table*}[]
\caption{Quantitative comparison for Light-LOAM \cite{10439642}, DLL \cite{9636501}, FAST-LIO2 \cite{9697912}, and Point-LIO \cite{Point-LIO}, and our method on three datasets (corresponding to subfigures (d), (g), and (j) in Fig. \ref{fig:15_images})}
\label{tab:my-table}
\resizebox{\textwidth}{!}{%
\begin{tabular}{@{}cccccccc@{}}
\toprule
Dataset &
  Case &
  Max absolute pose error(m) &
  Mean absolute pose error(m) &
  Max lateral error(m) &
  Mean lateral error(m) &
  Max longitudinal error(m) &
  Mean longitudinal error(m) \\ \midrule
\multirow{5}{*}{Fig. 4 (d)} & DLL        & 451.693 & 207.601 & 231.120 & 63.017  & 444.835 & 100.318 \\
                            & FAST-LIO2  & 8.229   & 2.753   & 8.053   & 1.879   & 2.845   & 0.274   \\
                            & Light-LOAM & 182.996 & 149.814 & 96.159  & 68.088  & 176.481 & 130.285 \\
                            & Point-LIO  & 55.524  & 3.590   & 5.643   & 1.563   & 55.236  & 1.131   \\
                            & Ours       & 0.742   & 0.165   & 0.236   & 0.031   & 0.704   & 0.030   \\ \midrule
\multirow{5}{*}{Fig. 4 (g)} & DLL        & 444.765 & 174.058 & 430.615 & 122.649 & 403.745 & 43.740  \\
                            & FAST-LIO2  & 10.311  & 3.922   & 9.819   & 1.490   & 10.263  & 1.657   \\
                            & Light-LOAM & 20.374  & 17.793  & 20.029  & 11.800  & 20.366  & 8.228   \\
                            & Point-LIO  & 39.408  & 20.741  & 36.930  & 14.726  & 38.253  & 4.734   \\
                            & Ours       & 0.927   & 0.200   & 0.898   & 0.066   & 0.533   & 0.025   \\ \midrule
\multirow{5}{*}{Fig. 4 (j)} & DLL        & 140.157 & 68.741  & 133.230 & 22.116  & 139.585 & 15.966  \\
                            & FAST-LIO2  & 2.161   & 0.921   & 1.290   & 0.404   & 2.158   & 0.169   \\
                            & Light-LOAM & 21.753  & 10.929  & 0.063   & 0.013   & 21.753  & 4.573   \\
                            & Point-LIO  & 17.937  & 3.934   & 7.833   & 0.433   & 16.435  & 1.680   \\
                            & Ours       & 0.553   & 0.262   & 0.511   & 0.116   & 0.472   & 0.075   \\ \bottomrule
\end{tabular}%
}
\vspace{-2em}
\end{table*}

\textbf{Implementation Platform and Hardware}.
Our dataset evaluation and comparison experiments are implemented in C++ using the Robot Operating System (ROS). The experiments are conducted on a computer equipped with an Intel i7-8750H CPU running at 2.20 G\,Hz, 32 GB of RAM, a GeForce GTX 1050 Ti GPU, and Ubuntu 20.04 as the operating system.

\textbf{Global Prior Map}. The port environment is highly dynamic, characterized by large IGVs, moving containers, massive gantry cranes, and rail cranes in constant motion. This ever-changing nature causes most environmental features to vary over time. Moreover, due to the port's unbounded and expansive layout, LiDAR data predominantly captures ground-level information. To create a stable and static global prior map, we conduct extensive validation and fine-tuning in real-world scenarios. During the map construction process, we prioritize generating a point cloud map that primarily focuses on ground features while systematically removing all potentially dynamic objects. This approach ensures reliable and consistent pose constraints for the localization algorithm. The final map covers an area of approximately one million square meters as shown in Fig.~\ref{fig:three-images}. 

\subsection{Localization Results and Comparison}
\label{sec:loc_result}



\textbf{Localization Results}.
Through testing and validation, it is observed that Light-LOAM \cite{10439642} and DLL \cite{9636501} fail to operate effectively across all port test datasets, showing significant errors. Consequently, the following analysis focuses on the algorithms that functioned properly: our algorithm, FAST-LIO2 \cite{9697912}, and Point-LIO \cite{Point-LIO}. The complete test results are shown in Fig.~\ref{fig:15_images}.

\textbf{Overall Trajectory Comparison.} The trajectory generated by our algorithm closely aligns with the ground truth (as shown in subfigures (a), (d), (g), (j), and (m) in Fig. \ref{fig:15_images}), with near-complete overlap, demonstrating exceptional accuracy in global localization. In contrast, on more complex paths (e.g., subfigures (a), (c), and (e) in Fig. \ref{fig:15_images}), the trajectories of FAST-LIO2 \cite{9697912} and Point-LIO \cite{Point-LIO} exhibit varying degrees of deviation. Notably, Point-LIO \cite{Point-LIO} shows more significant errors, especially in sharp turns or sections with complex motion, revealing its reduced robustness in dynamic scenarios. Our algorithm, however, maintains superior accuracy even under these challenging conditions, with minimal trajectory deviation, showcasing its strong adaptability and reliability.

\textbf{Translational Comparison.} The analysis of the translational component, examining accuracy in the $x$, $y$, and $z$ directions, is presented in subfigures (b), (e), (h), (k), and (n) in Fig. \ref{fig:15_images}. In the $x$ and $y$ directions, all algorithms perform well, with their estimated curves closely matching the ground truth, reflecting high translational accuracy. However, in the $z$ direction (height), FAST-LIO2 \cite{9697912} and Point-LIO \cite{Point-LIO} show noticeable deviations, as observed in subfigures (f), (h), and (j) in Fig. \ref{fig:15_images}. In contrast, our algorithm exhibits significantly better performance with smaller deviations. This discrepancy may stem from substantial sensor noise in height data or inadequate optimization of height estimation in the competing algorithms.

\textbf{Rotation Comparison.} The rotational component analysis, focusing on angle estimation around the $x$, $y$, and $z$ axes, places particular emphasis on the yaw angle, as shown in subfigures (c), (f), (i), (l), and (o) in Fig. \ref{fig:15_images}. Our algorithm demonstrates superior rotational accuracy across all datasets, with its estimated curves closely following the ground truth. While slight jitter is observed, it remains minimal and does not impact overall performance. In contrast, Point-LIO \cite{Point-LIO} and FAST-LIO2 \cite{9697912} exhibit significant yaw estimation jitter in certain datasets (e.g., subfigures (k) and (o) in Fig. \ref{fig:15_images}), especially near the end of the tests (the far-right sections of the figures), where error magnitudes increase considerably. This behavior is likely due to the inability of these algorithms to effectively correct accumulated angular errors.

\textbf{Quantitative Comparison.} We evaluated the performance of DLL \cite{9636501}, FAST-LIO2 \cite{9697912}, Light-LOAM \cite{10439642}, Point-LIO \cite{Point-LIO}, and our approach on three datasets (Fig. \ref{fig:15_images} (d), (g), and (j)) using metrics such as maximum and mean absolute pose errors, lateral errors, and longitudinal errors. The results in TABLE \ref{tab:my-table}, consistently show that our method outperforms all other algorithms across all datasets and metrics. In Fig. \ref{fig:15_images} (d), our method achieved a maximum absolute pose error of 0.742\,m, significantly lower than DLL \cite{9636501} (451.693\,m). Its mean absolute pose error was 0.165\,m, far below FAST-LIO2 \cite{9697912} (2.753\,m) and Point-LIO \cite{Point-LIO} (3.590\,m). Similarly, in Fig. \ref{fig:15_images} (g), our system maintained the smallest maximum (0.927\,m) and mean (0.200\,m) pose errors, with minimal lateral and longitudinal errors. In Fig. \ref{fig:15_images} (j), our algorithm continued to excel, achieving a maximum pose error of 0.553\,m and a mean pose error of 0.262 m, both significantly better than the competing algorithms. Besides, DLL \cite{9636501} and Light-LOAM \cite{10439642} fail with large errors.

Overall, our framework consistently demonstrated the highest accuracy and robustness, with significantly lower errors across all datasets. These results highlight the effectiveness of its improved optimization strategies and robust error-handling mechanisms in complex environments.


\section{CONCLUSION}

In this paper, we present a robust and precise mapping and localization system designed for changing, dynamic outdoor environments. By tightly coupling a LiDAR-inertial odometry with map-based localization, the system addresses instability and drift caused by environmental changes, integrating offline global map registration with real-time dynamic mapping for reliable performance in large-scale scenarios. At its core, the system employs a multi-sensor fusion framework based on the iESKF, ensuring both robustness and accuracy. Extensive real-world experiments demonstrate that our method consistently outperforms state-of-the-art LiDAR localization algorithms, making it a reliable solution for applications like robotics and autonomous driving.

\bibliographystyle{IEEEtran}
\bibliography{IEEEabrv,IEEEexample,mybibfile}

\end{document}